%% file: bmvc_final.tex
\documentclass{bmvc2k}

\usepackage{times}
\usepackage{epsfig}
\usepackage{graphicx}
\usepackage{amsmath}
\usepackage{amssymb}

\usepackage{multirow}
\usepackage{makecell}
\usepackage{algorithm}
\usepackage{algorithmic}
\usepackage[para,online,flushleft]{threeparttable}
\usepackage{adjustbox}

\usepackage{booktabs}
\usepackage{wrapfig,lipsum,booktabs}

\title{Semantic Adversarial Attacks via\\Diffusion Models}

\addauthor{Chenan Wang}{cw3344@drexel.edu}{1}
\addauthor{Jinhao Duan}{jd3734@drexel.edu}{1}
\addauthor{Chaowei Xiao}{xiaocw@umich.edu}{2}
\addauthor{Edward Kim}{ek826@drexel.edu}{1}
\addauthor{Matthew Stamm}{mcs382@drexel.edu}{3}
\addauthor{Kaidi Xu}{kx46@drexel.edu}{1}
\addinstitution{
Department of Computer Science\\
College of Computing \& Informatics\\
Drexel University\\
 Philadelphia, USA
}
\addinstitution{
The Information School\\
University of Wisconsin\\
Madison, USA
}
\addinstitution{
Electrical and Computer Engineering\\
Drexel University\\
Philadelphia, USA
}

\runninghead{Wang et al.}{Semantic Adversarial Attacks via Diffusion Models}

\begin{document}

\maketitle

\begin{abstract}
Traditional adversarial attacks concentrate on manipulating clean examples in the pixel space by adding adversarial perturbations. By contrast, semantic adversarial attacks focus on changing semantic attributes of clean examples, such as color, context, and features, which are more feasible in the real world. In this paper, we propose a framework to quickly generate a semantic adversarial attack by leveraging recent diffusion models since semantic information is included in the latent space of well-trained diffusion models. Then there are two variants of this framework: 1) the \textbf{S}emantic \textbf{T}ransformation (ST) approach fine-tunes the latent space of the generated image and/or the diffusion model itself; 2) the \textbf{L}atent \textbf{M}asking (LM) approach masks the latent space with another target image and local backpropagation-based interpretation methods. Additionally, the ST approach can be applied in either white-box or black-box settings. Extensive experiments are conducted on CelebA-HQ and AFHQ datasets, and our framework demonstrates great fidelity, generalizability, and transferability compared to other baselines. 
Our approaches achieve $\sim$100\% attack success rate in multiple settings with the best FID as 36.61. Code is available at \url{https://github.com/steven202/semantic_adv_via_dm}.

\end{abstract}

\section{Introduction}

Deep neural networks have achieved breakthroughs in many domains~\cite{he2016deep,yuan2020attribute,yuan2023remind,bau2020understanding}, however, their intrinsic vulnerabilities to adversarial examples raise security concerns~\cite{carlini2017towards,zhao2019admm,xu2020adversarial,zhang2022branch}. Most of the literature on adversarial machine learning has been generalized to adversarial perturbations within a $\ell_p$ norm ball with a small radius $\epsilon$ around the clean input example. Vanilla-trained models achieve high accuracy in classifying benign examples, while misclassifying inputs with such imperceptible perturbations. 
Instead of globally attacking input images on pixel space, \cite{hosseini2018semantic,song2018constructing} proposed semantic adversarial attacks gaining insight into real-world robustness by manipulating semantically meaningful visual attributes. 
Semantic attacks may be perceptible; however, such attacks are semantically meaningful and thus hard to detect. 
Following the concept of semantic adversarial attacks, there is a growing literature on this topic~\cite{joshi2019semantic,bhattad2019unrestricted,qiu2020semanticadv,shamsabadi2020colorfool,xu2020towards,wang2021demiguise}. In the real world, adversarial attacks in $\ell_p$-norm based constraint rarely happen due to fragile perturbations. 
Compared with $\ell_p$-norm adversarial attacks in the pixel space,  semantic adversarial images are more feasible since they are unrestricted in the magnitude of perturbation while preserving perceptual similarity and realism. Such attacks include changes in texture or any semantic attribute that lead to misclassification.

\begin{wrapfigure}{r}{7cm}
\vspace{-5mm}
\centering
\renewcommand*{\arraystretch}{0.0}
\begin{tabular}{>{\centering}m{0.04\textwidth}  >{}m{0.12\textwidth} >{}m{0.12\textwidth} >{}m{0.12\textwidth} }
& \thead{Original\\image} & \thead{Difference} & \thead{Adv.\\image}
\\
{\thead{white-\\box}}
& 
\includegraphics[width=0.4\textwidth]{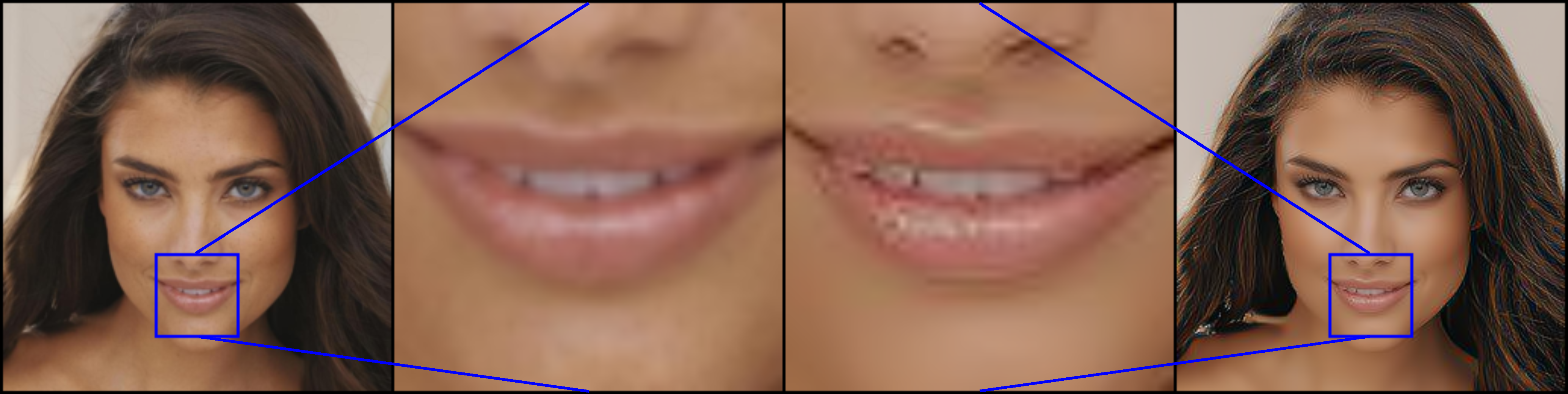}
\\
{\thead{black-\\box}}
& 
\includegraphics[width=0.4\textwidth]{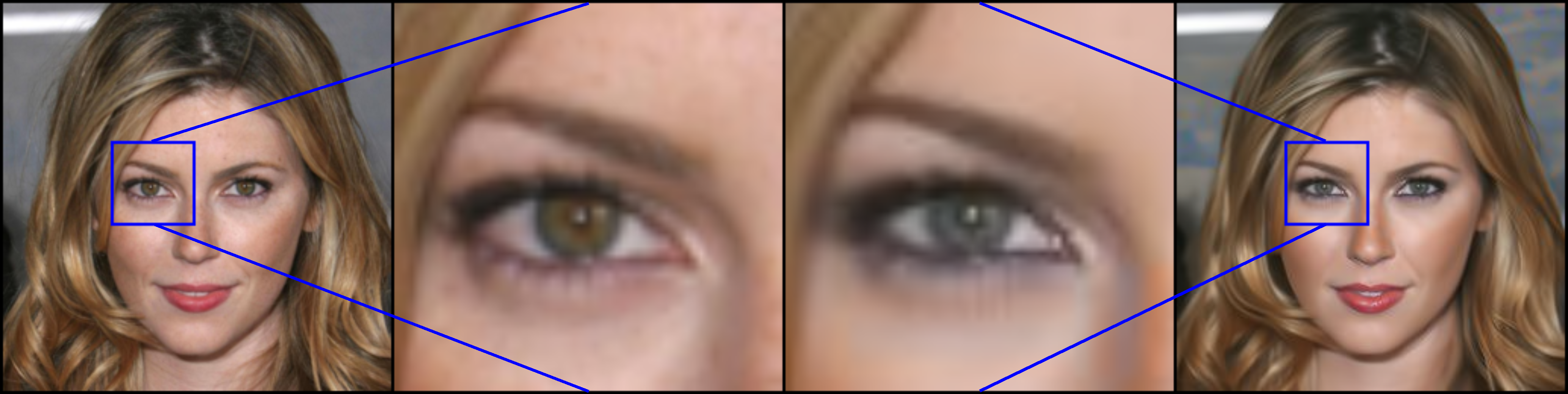}
\end{tabular}
\begin{tabular}{>{\centering}m{0.04\textwidth}>{}m{0.08\textwidth}>{}m{0.08\textwidth}>{}m{0.06\textwidth}>{}m{0.10\textwidth}}
& \thead{Source\\image} & \thead{Target\\image} &  \thead{Mask} & \thead{Adv. \, \\image \,}
\\
{\thead{\, Grad-\\ \, CAM}} & 
\includegraphics[width=0.4\textwidth]{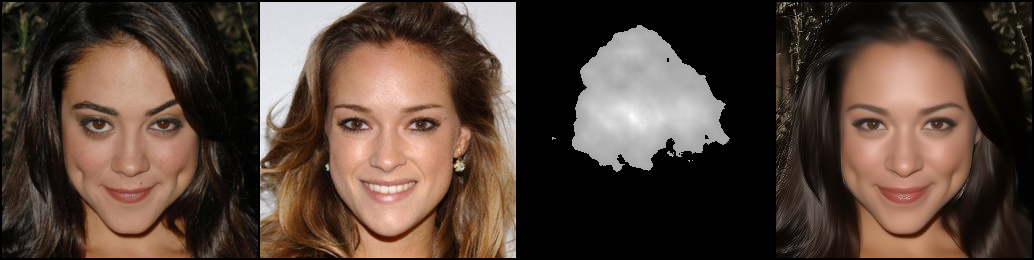}
\\
{\thead{saliency\\maps}}
& 
\includegraphics[width=0.4\textwidth]{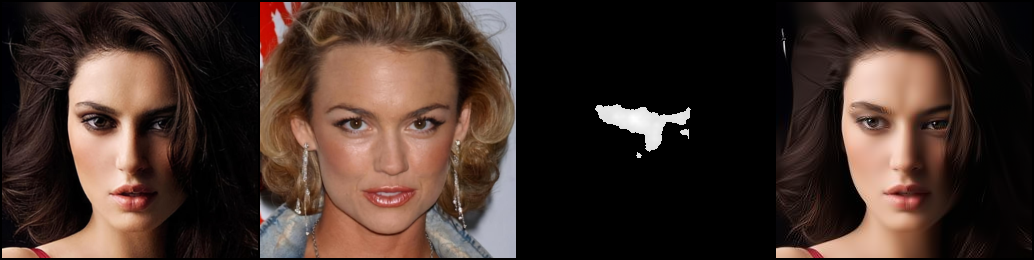}
\end{tabular}
\vspace{4mm}
\caption{
Top two rows: our framework with ST approach under white-box and black-box settings with small semantic changes.
Bottom two rows: our framework with LM approach masking by Grad-CAM and saliency maps to transplant features from the target image. 
\label{fig:intro}
}
\vspace{-2mm}
\end{wrapfigure}

Currently, there are two existing approaches to implement semantic attacks on clean images: transformations in color or texture~\cite{hosseini2018semantic,bhattad2019unrestricted}, or by performing manipulation in the latent space of a generative model~\cite{joshi2019semantic,qiu2020semanticadv}, such as Generative Adversarial Networks (GANs)~\cite{goodfellow2020generative}. 
The former leverages various techniques to gather color and texture information, while the latter relies upon attribute annotations to generate semantic adversarial images with generative models such as GANs. 
Most previous works on semantic adversarial attacks utilize generative models to change attributes, relying on attribute annotations~\cite{joshi2019semantic,qiu2020semanticadv}, color or texture information~\cite{bhattad2019unrestricted,hosseini2018semantic,shamsabadi2020colorfool}.

While previous approaches have demonstrated the feasibility of semantic attacks, images that have been attacked using these algorithms can often be easily spotted by the human eye. In order for a semantic attack to be successful, the attacked image should not only fool the classifier but also appear convincingly realistic. Furthermore, existing approaches take a significant time to generate a single attacked image.  As a result, it is not feasible to launch these attacks at scale. To address these problems, we develop our framework by leveraging diffusion models (DMs) without any other annotations. Recently, DMs have drawn significant attention in the image generation area with higher fidelity~\cite{song2020score,ho2020denoising} over GANs on image synthesis~\cite{dhariwal2021diffusion}. Also, the latent space in a DM intrinsically contains semantic information.
Similar to other generative models, DMs also provide a latent space, but because it is in the same dimensions as input and output, attack methods can easily map the features from the latent space to the generated image. 

With exploiting DMs, in this paper, we propose the \textbf{S}emantic \textbf{T}ransformation (ST) approach, which requires gradient information from the target classifier in the white box setting or leverages a surrogate model in the black box setting to generate minimal semantic changes on the original inputs. Also, we provide another variant, the \textbf{L}atent \textbf{M}asking (LM) approach, that can transplant auto-selected features by different masking methods from a target image. Some visual examples are shown in Figure~\ref{fig:intro}.  We summarize our main contributions as follows:
\vspace{-1mm}
\begin{itemize}
  \setlength\itemsep{0.1em}
    \item We propose the \textbf{S}emantic \textbf{T}ransformation (ST) approach, either white-box or black-box, a generalized way to generate semantic adversarial attacks via fine-tuning a latent space and/or a diffusion model. 
    \item We propose the \textbf{L}atent \textbf{M}asking (LM) approach to fast generate semantic adversarial attacks. Two interpretation methods (Grad-CAM or saliency map) are used to mask a latent space by transplanting another target image.
    \item Our framework with ST and LM approaches is the first systematic way to generate semantic adversarial attacks leveraging diffusion models. 
\end{itemize}

\paragraph{Related Work} 
Unlike $\ell_p$-norm based attacks with pixel-wise perturbations, \cite{xiao2018spatially,dong2018boosting,brown2018unrestricted} proposed unrestricted adversarial attacks with other techniques such as spatial transformations. At the same time, \cite{hosseini2018semantic} generate adversarial examples by manipulating the colors of a clean image in the HSV color space, such as randomly shifting the hue and saturation components. Concurrently, \cite{song2018constructing} proposed a non $\ell_p$-norm based attack, generated with conditional generative models. Even the concept of semantic attack was not mentioned in this work, since the attacks are generated with semantic information, it proposed the prototype of semantic attack: a kind of unrestricted perturbations with $\ell_p$-norm by manipulating semantic information while keeping perceptual similarity realism. As aforementioned, changing color or texture as in~\cite{hosseini2018semantic,bhattad2019unrestricted}, and manipulating attributes by generative models are both considered semantic attacks, and many studies~\cite{joshi2019semantic,bhattad2019unrestricted,qiu2020semanticadv,shamsabadi2020colorfool,xu2020towards,wang2021demiguise} built on top of them. Most of such attacks are visible; such as~\cite{qiu2020semanticadv}, where it proposed a human face-based semantic attack algorithm by slightly changing the attributes (e.g.,  with the additional annotation. Though some~\cite{wang2021demiguise,na2022unrestricted,xu2020towards} of them are not visible, it is still a semantic attack, as long as it uses semantic information to modify a clean image. For example, \cite{wang2021demiguise} crafted invisible semantic adversarial perturbations by manipulating semantic information with Perceptual Similarity (PS), and \cite{na2022unrestricted} generate targeted unrestricted adversarial attacks with a decision-based attacking algorithm in a latent space of an adversarial generative model (GAN).
However, when generating semantic adversarial attacks with generative models, previous studies either rely on a dataset with attribute annotations~\cite{qiu2020semanticadv} or require thousands of queries~\cite{na2022unrestricted}. 
Traditional image generation techniques usually require adversarial generative networks (GANs); recently diffusion models (DMs)~\cite{ho2020denoising,song2020score,duan2023diffusion} achieved superior image quality to GANs on image synthesis~\cite{dhariwal2021diffusion}.

\section{Methodology}
\subsection{Preliminary: Diffusion Models}
Diffusion models~\cite{ho2020denoising,song2020score} include a diffusion process (forward process) and a sampling process (reverse process). The diffusion process transforms data to a simple noise distribution while the sampling process reverses this process. Either of the two steps is a Markov chain and consists of a sequence of steps, where every step can be approximated to a Gaussian distribution. The diffusion and sampling processes
can be defined as follows:
\begin{equation}\label{eq:diffusion_process}
\vspace{-2mm}
{\small
\begin{aligned}
& q_\theta(\mathbf{x}_{1:T}|\mathbf{x}_{0})= \prod_{t=1}^{T}q_\theta (\mathbf{x}_{t}|\mathbf{x}_{t-1}),\, q (\mathbf{x}_{t}|\mathbf{x}_{t-1})=\mathcal{N}(\mathbf{x}_{t};\sqrt{1-\beta_t}\mathbf{x}_{t-1},\beta_t\mathbf{I}),
\end{aligned}
}
\end{equation}
\begin{equation}\label{eq:sampling_process}
{\small
\begin{aligned}
& p_\theta(\mathbf{x}_{0:T})=  p(\mathbf{x}_T)\prod_{t=1}^{T}p_\theta (\mathbf{x}_{t-1}|\mathbf{x}_t),\, p_\theta (\mathbf{x}_{t-1}|\mathbf{x}_t)=\mathcal{N}(\mathbf{x}_{t-1};\mathbf{\mu}_\theta(\mathbf{x}_t,t),\Sigma_\theta(\mathbf{x}_t,t)),
\end{aligned}
}
\end{equation}
where $\mathbf{x}_t$ is the latent space for $t=1,\cdots,T$. The latent space $\mathbf{x}_t$ in diffusion process can be expressed as:
\begin{equation}\label{eq:generated_latent}
\begin{aligned}
& \mathbf{x}_t=\sqrt{\bar{\alpha}_t}\mathbf{x}_0+\sqrt{1-\bar{\alpha}_t}\mathbf{w},\;
\mathbf{w}\sim \mathcal{N}(\mathbf{0},\mathbf{I}),
\end{aligned}
\end{equation}
where $\alpha_t= 1-\beta_t$ and $\bar{\alpha}_t = \prod_{s=1}^{t}\alpha_s$.
In the forward process, the parameter $\{{\beta_t}\}_{t=0}^T$ can be either a learnable parameter by reparameterization~\cite{ho2020denoising} or fixed constant. In the reverse process, $\mathbf{\mu}_\theta(\mathbf{x}_t,t)$ can be expressed as $\mathbf{\mu}_\theta(\mathbf{x}_t,t)= \frac{1}{\sqrt{\alpha_t}}(\mathbf{x}_t-\frac{\beta_t}{\sqrt{1-\bar{\alpha}_t}}\mathbf{\epsilon}_\theta(\mathbf{x}_t,t))$,
where $\mathbf{\epsilon}_\theta(\mathbf{x}_t,t)$ is a noise approximation model, which predicts $\mathbf{\epsilon}$ from a latent space $\mathbf{x}_t$ and a time step $t$.
This model can be trained by minimizing the following loss function over model parameters $\theta$ as $\mathcal{L}_{simple}(\theta)=\mathbb{E}_{x_0\sim q(\mathbf{x}_0),\mathbf{w}\sim\mathcal{N}(\mathbf{0},\mathbf{I}),t}{\lVert \mathbf{w}-\mathbf{\epsilon}_\theta(\mathbf{x}_t,t) \rVert}^2_2.$

After the model is trained, data can be sampled with the sampling process as $\mathbf{x}_{t-1}=\frac{1}{\sqrt{\alpha_t}}(\mathbf{x}_t-\frac{1-\alpha_t}{\sqrt{1-\bar{\alpha}_t}}\mathbf{\epsilon}_\theta(\mathbf{x}_t,t))+\sigma_t\mathbf{z}$,
where $\mathbf{z}\sim\mathcal{N}(\mathbf{0},\mathbf{I})$. 
Meantime,~\cite{song2020denoising} proposed a non-Markovian diffusion process that leverages the same DDPM in the forward process as they share the same forward marginals, and it has a distinct sampling process:
\begin{equation}\label{eq:generated_image}
\vspace{-2mm}
\begin{aligned}
\mathbf{x}_{t-1}=\sqrt{\bar{\alpha}_{t-1}}\mathbf{f}_\theta(\mathbf{x}_t,t)+\sqrt{1-\bar{a}_{t-1}-\sigma_t^2}\mathbf{\epsilon}_\theta(\mathbf{x}_t,t)+\sigma_t^2\mathbf{z},
\end{aligned}
\end{equation}
where $\mathbf{z}\sim\mathcal{N}(\mathbf{0},\mathbf{I})$ and $\mathbf{f}_\theta(\mathbf{x}_t,t)$ is a estimation of $\mathbf{x}_0$ at t given $\mathbf{x}_t$ and $\mathbf{\epsilon}_\theta(\mathbf{x}_t,t)$, as $\mathbf{f}_\theta(\mathbf{x}_t,t)= \frac{\mathbf{x}_t-\sqrt{1-\bar{\alpha}_t}\mathbf{\epsilon}_\theta(\mathbf{x}_t,t)}{\sqrt{\bar{\alpha}_t}}$.
\subsection{The \textbf{S}emantic \textbf{T}ransformation (ST) approach}
\begin{figure}[!ht]  
  \includegraphics[width=0.97\textwidth]{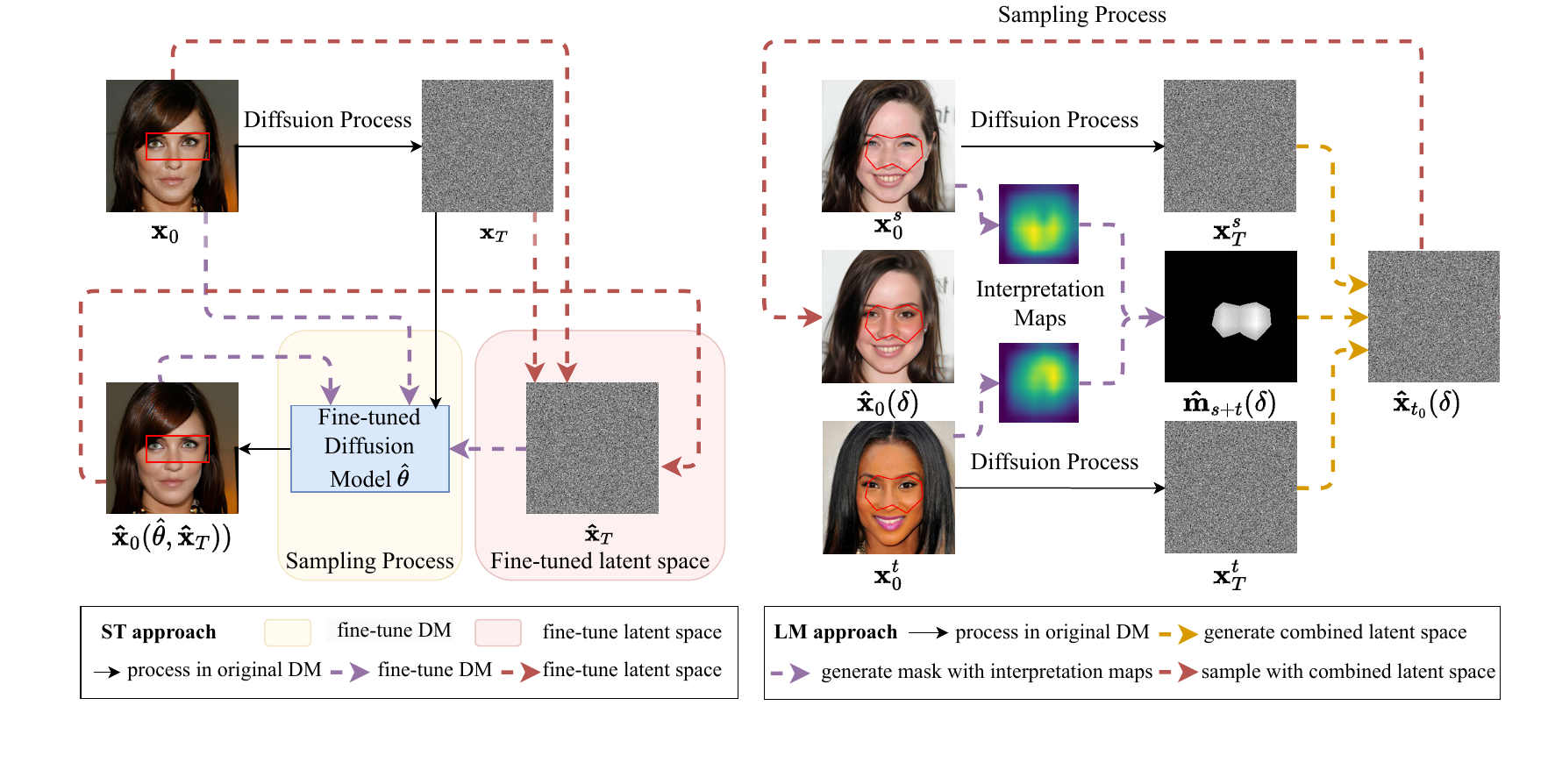}
  \vspace{-5mm}
    \caption{
    Pipelines of generating semantic adversarial images via a fine-tuning process by ST and LM approach.}
      \label{fig:pipeline_st_lm}
    \vspace{-4mm}
\end{figure}

One crude way of using a diffusion model to perform a semantic adversarial attack is to directly manipulate the latent space through an attack loss as shown in Figure~\ref{fig:pipeline_st_lm}.
Given a clean image $\mathbf{x}_0$, we use Eq.\eqref{eq:generated_latent} to obtain its latent space $\mathbf{x}_{T}$ with a diffusion process as in Eq.~\eqref{eq:diffusion_process}. The generated semantic adversarial image is denoted as $\mathbf{\hat{x}}_0(\hat{\theta}, \mathbf{\hat{x}}_{T})$ with the fine-tuned diffusion model parameter $\hat{\theta}$ and the fine-tuned latent space $\mathbf{\hat{x}}_{T}$. We fine-tune the latent space and/or the diffusion model to transform semantic information during the fine-tuning process until the generated image $\mathbf{\hat{x}}_0(\hat{\theta}, \mathbf{\hat{x}}_{T})$ mislead the classifier. After the fine-tuning process, we evaluate our attacks with a sampling process as in Eq.~\eqref{eq:sampling_process}. For diffusion and sampling processes, we use DDIM~\cite{song2020denoising} as it is a deterministic process. The fine-tuning process is performed until the target classifier can be fooled by generated image $\mathbf{\hat{x}}_0(\hat{\theta}, \mathbf{\hat{x}}_{T})$. As shown in \cite{kwon2022diffusion}, we believe manipulating a latent space $\mathbf{x}_{T}$ at step $T$ affects the generated image after sampling, since $\mathbf{x}_{T}$ contains semantic information of the original image $\mathbf{x}_0$. The algorithm is in the supplementary material.

\vspace{1mm}
\noindent\textbf{Loss Function in the Finetuning Process.}
There are many perceptual metrics for assessing the perceptual similarity between two images, such as Peak signal to noise ratio (PSNR) and structural index similarity (SSIM); however, these metrics fail to capture the nuances of human perceptions~\cite{zhang2018unreasonable}. Evaluated on BAPPS dataset, \cite{zhang2018unreasonable} proposed the Learned Perceptual Image Patch Similarity (LPIPS) metric, which recognizes similarities well even with various distortions for a pair of images. 
Hence, we minimize the LPIPS metric in our loss function, to maximize the perceptual similarity between the original image and the generated semantic adversarial image. 
Inspired by the TRADES loss~\cite{zhang2019theoretically}, we maximize the KL divergence between the prediction logits on the original image and the prediction logits on the semantic adversarial image. Our loss function in the fine-tuning process is defined as: 

\begin{equation}\label{eq:st_loss}\vspace{-2mm}
\begin{aligned}[t]
\mathcal{L}_{\textit{ST}} =\min_{\hat{\theta}, \mathbf{\hat{x}}_{T}}\, &\lambda \mathit{D}_{\textit{LPIPS}}(\mathbf{x}_0, \mathbf{\hat{x}}_0(\hat{\theta}, \mathbf{\hat{x}}_{T}))-\mathit{D}_{\textit{KL}}(f(\mathbf{x}_0), f(\mathbf{\hat{x}}_0(\hat{\theta}, \mathbf{\hat{x}}_{T}))),
\end{aligned}
\vspace{-1mm}
\end{equation}
where $\mathit{D}_{\textit{LPIPS}}$ captures the perceptual similarities of the original image $\mathbf{x}_0$ and the generated adversarial image $\mathbf{\hat{x}}_0(\hat{\theta}, \mathbf{\hat{x}}_{T}))$, and minimizing this loss term keeps perceptual features of the pair of images remain same during fine-tuning. By contrast, maximizing $\mathit{D}_{\textit{KL}}$ encourages the generated image $\mathbf{\hat{x}}_0(\hat{\theta}, \mathbf{\hat{x}}_{T}))$ to enlarge the logits distance with $\mathbf{x}_0$ with respect to a classifier $f$, either a known classifier or a random pre-trained classifier. Thus, there is a trade-off between these two terms: while maintaining the global perceptual similarities, we expect to change the local attributes misleading the classifier $f$. The relative strengths of two loss terms $\mathit{D}_{\textit{LPIPS}}$ and $\mathit{D}_{\textit{KL}}$ can be adjusted by the scalar $\lambda$. 
 
\vspace{1mm}
\noindent\textbf{Nuance between White-box and Black-box Attacks.}
The difference between a white-box and black-box attack is whether the malicious knows the target model parameters of~$f$. 
In Eq.~\eqref{eq:st_loss}, when calculating  $\mathit{D}_{\textit{KL}}$, the prediction logits from the target classifier are used for a white-box attack, and the outputs from a pre-trained InceptionV3 model~\cite{szegedy2015rethinking} are used for a black-box attack. 

\begin{figure}
\centering
\renewcommand*{\arraystretch}{1.0}
\begin{tabular}{*{8}{p{1.125cm}}}
\small{Original} & \small{Enlarged} & \small{Enlarged} & \small{Attack} & \small{Original} & \small{Enlarged} & \small{Enlarged} & \small{Attack}\\
\multicolumn{4}{c}{      \includegraphics[width=0.450\textwidth]{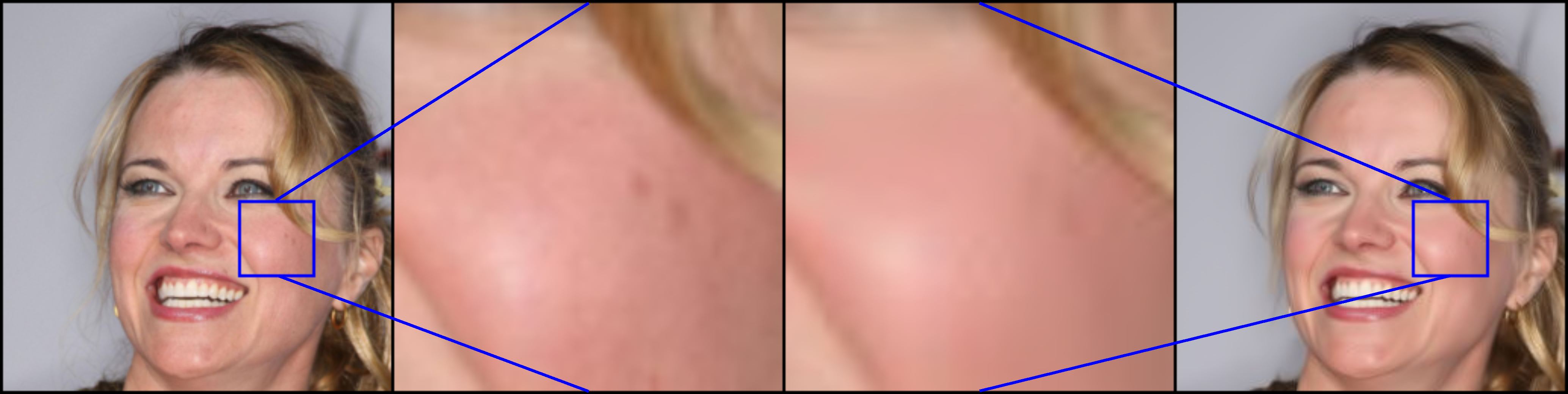}
}&
\multicolumn{4}{c}{\includegraphics[width=0.450\textwidth]{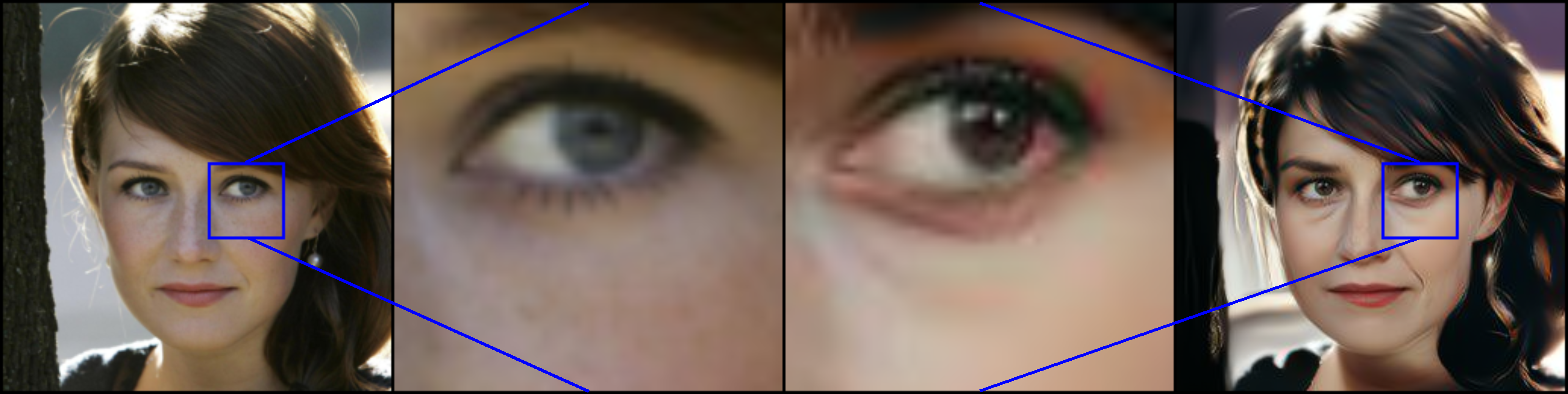}}
\\
\multicolumn{4}{c}{      \includegraphics[width=0.450\textwidth]{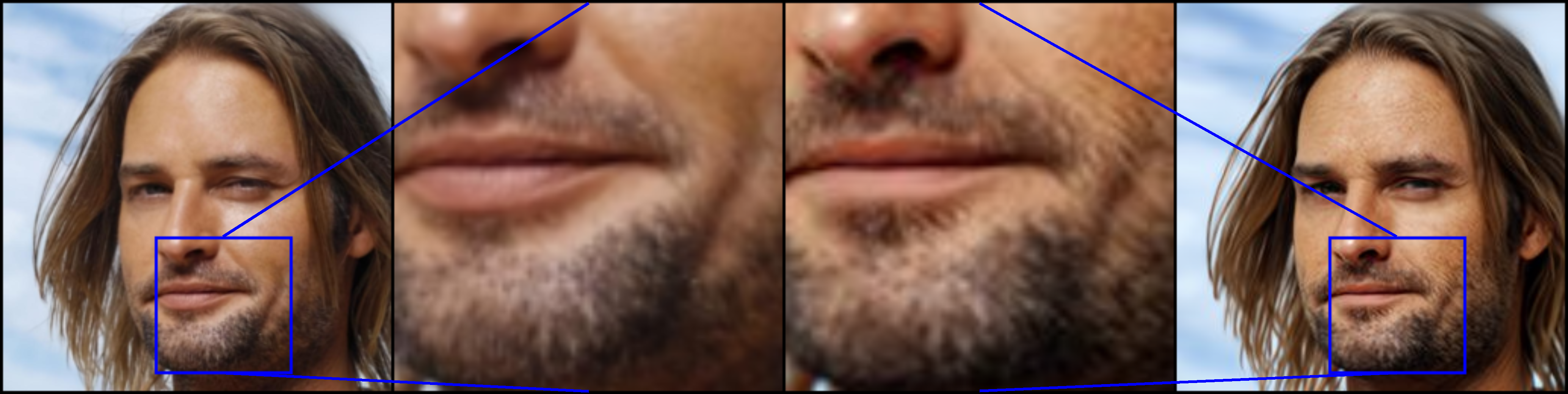}
}&
\multicolumn{4}{c}{    \includegraphics[width=0.450\textwidth]{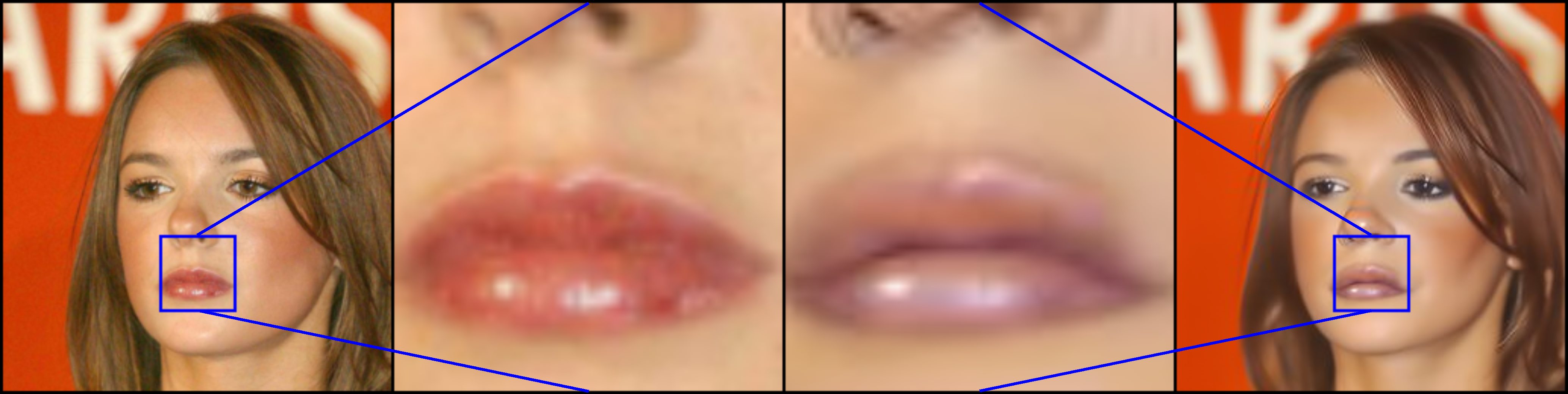}
}
\\
\multicolumn{4}{c}{      \includegraphics[width=0.450\textwidth]{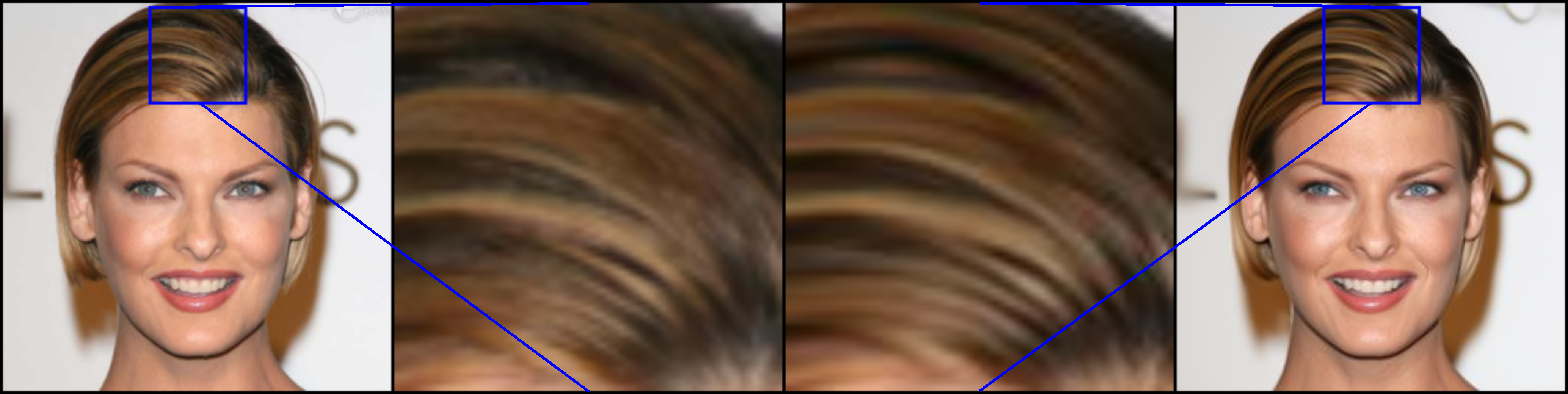}
}&
\multicolumn{4}{c}{      \includegraphics[width=0.450\textwidth]{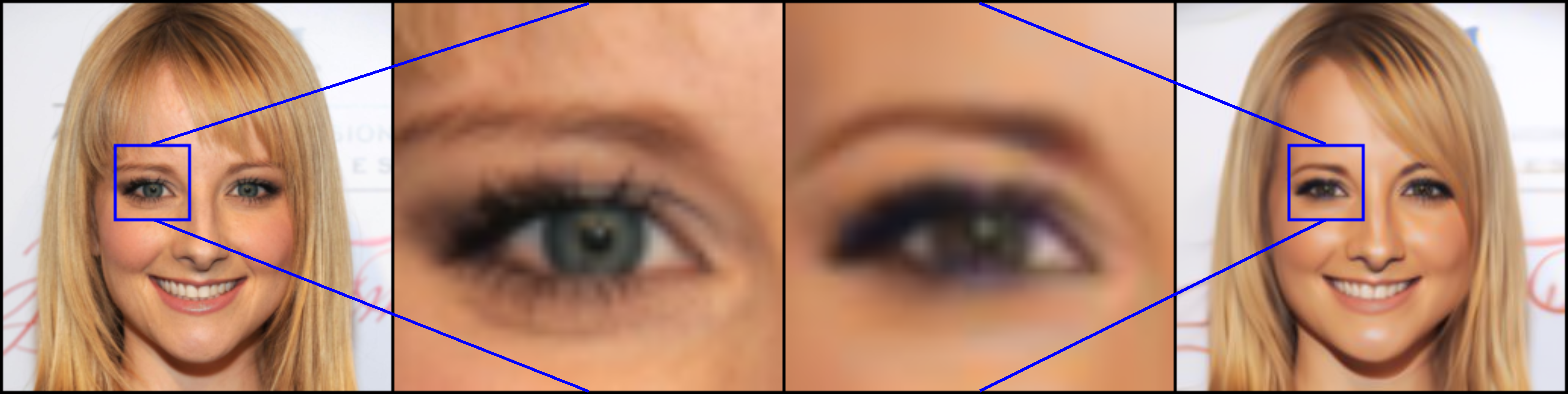}
}
\\
\multicolumn{4}{c}{White-box Attack}&
\multicolumn{4}{c}{Black-box Attack}
\\
\end{tabular}
\vspace{2mm}
\caption{
ST approach under different fine-tuning. First row: only fine-tuning a latent space. Second row: only fine-tuning a diffusion model. Third row: fine-tuning both a latent space and a diffusion model. First column: original image. Second column: enlarged local area from the original image. Third column: enlarged local area from the adversarial image. Fourth column: adversarial image.}
\label{fig:st}
\vspace{-3mm}
\end{figure}

\vspace{-2mm}
\subsection{The \textbf{L}atent \textbf{M}asking (LM) Approach}
Fine-tuning a latent space or diffusion model requires calculating the gradients on the latent space or diffusion model parameters, resulting in huge computation expenses. In this subsection, we introduce an alternative approach to modify the generated image, that is, by masking the latent space with feature significance from a target image. The mask area is the most significant in a feature map and is intended to contain important, semantically meaningful features with respect to the target classifier.  We transplant the masked area as in Figure~\ref{fig:pipeline_st_lm}. 
Let $\mathbf{m}$ be an interpretation map, it can be calculated as $\mathbf{m} = g(\mathbf{x}_0,\mathbf{y})$, where $\mathbf{x}_0$ is a clean image, and $\mathbf{y}$ is its label. We use Grad-CAM~\cite{selvaraju2016grad,jacobgilpytorchcam} and saliency maps~\cite{simonyan2013deep,srinivas2019full} for interpretation maps in this paper. 

\vspace{1mm}
\noindent\textbf{Transplanting Features with Mask.}
We denote a pair of source and target images as $\mathbf{x}^s_0$ and $\mathbf{x}^t_0$, and their latent spaces as $\mathbf{x}^s_T$ and $\mathbf{x}^t_T$ respectively. Here, $\mathbf{x}^s_0$ is used as the victim image, and most features in $\mathbf{x}^s_0$ 
are kept; $\mathbf{x}^t_0$ is used as the target image, and a small portion of masked features in $\mathbf{x}^t_0$ 
are transplanted to $\mathbf{x}^s_0$ when generating a semantic adversarial attack. We use Grad-CAM or saliency map to generate the mask. The significance maps are calculated on $\mathbf{x}^s_0$ and $\mathbf{x}^t_0$, denoted as $\mathbf{m}_s$ and $\mathbf{m}_t$. We have three strategies to generate a mask $\mathbf{\hat{m}}$ and we denoted them as $\mathbf{\hat{m}}_{s}(\delta)$, $\mathbf{\hat{m}}_{t}(\delta)$ and $\mathbf{\hat{m}}_{{s+t}}(\delta)$, respectively:
\begin{equation}\small\label{eq:lm_mask_generation}
\vspace{-2mm}
\begin{aligned}[t]
\mathbf{\hat{m}}_{s}(\delta) = \textit{TopK}(|\mathbf{m}_s|, \delta),\quad 
\mathbf{\hat{m}}_{t}(\delta) = \textit{TopK}(|\mathbf{m}_t|, \delta), \quad 
\mathbf{\hat{m}}_{{s+t}}(\delta) = \textit{TopK}(|\mathbf{m}_s|+|\mathbf{m}_t|, \delta),\\  
\end{aligned}
\end{equation}
\noindent
where $\delta$ is a percentage threshold of $\textit{TopK}$ function, ranging from 0 to 99, and $\textit{TopK}$ function would only keep a given input with the $\delta\%$ largest and set other elements in the mask as zero. 
We design a heuristic to control the decremental speed of $\delta$:
\begin{equation}\small\label{eq:delta_decremental_speed}
\vspace{-2mm}
\begin{aligned}[t]
\delta=\delta - \max(\gamma\frac{z_y - \max_{i\neq y}z_i}{z_y}, 1), 
\end{aligned}
\end{equation}
where $z_y$ is the target class confidence logit, $\max_{i\neq y}z_i$ is the second highest confidence logit, and $\gamma$ is a constant.
With the mask $\mathbf{\hat{m}}_{t_0}(\delta)$ created, the original latent space is modified as:
\begin{equation}\small\label{eq:lm_latent_generation}
\vspace{-2mm}
\begin{aligned}[t]
\mathbf{\hat{x}}_{T}(\delta) =&(1-\mathbf{\hat{m}}(\delta))*\mathbf{x}^s_T+\mathbf{\hat{m}}(\delta)*\mathbf{x}^t_T
\end{aligned}
\end{equation}
With modified latent space $\mathbf{\hat{x}}_{T}(\delta)$, we generated a semantic adversarial image without fine-tuning a latent space or diffusion model following Eq.\eqref{eq:generated_image}. The whole process of the LM approach is shown in Figure~\ref{fig:pipeline_st_lm}. The algorithm is in the supplementary material.
\vspace{-1mm}
\section{Experiments}
\vspace{-1mm}
\subsection{Datasets}
\label{sec:datasets}
\vspace{-1mm}
 We evaluate our white-box and black-box attacks on two tasks: human facial identity recognition and animal category recognition. For all experiments, we use 500 images with the size of 256 $\times$ 256.
 
\vspace{1mm}
\noindent\textbf{Human Facial Identity Recognition and Gender Classification.}
We use Celeb-HQ Facial Identity Recognition Dataset, which is a subset of the CelebAMask-HQ dataset~\cite{CelebAMask-HQ}, adopted from~\cite{na2022unrestricted}. The original CelebAMask-HQ dataset contains 30,000 face images at 512 x 512 resolution. It has 6,217 unique identities. For the target classifier, we use a subset of it as in~\cite{na2022unrestricted}, which contains 307 unique identities, 4,263 images for training, and 1,215 images for testing. A ResNet18~\cite{he2016deep} classifier is trained for 30 epochs with 89.05\% accuracy on this dataset. For the diffusion model, we used the model pretrained on CelebA-HQ~\cite{karras2017progressive} dataset, which contains 30,000 images at 1024 x 1024 resolution. The evaluation is in Section~\ref{sec:results} and the supplementary material. 
Besides facial identity recognition, we also adopt the Celeb-HQ Face Gender Recognition Dataset from~\cite{na2022unrestricted}, which contains 11,057 male and 18,943 female images, and the evaluation is in the supplementary material. 

\vspace{1mm}
\noindent\textbf{Animal Category Recognition.}
We use the AFHQ~\cite{choi2020starganv2} dataset, a dataset of animal faces consisting of 15,000 images at 512 x 512 resolution. The dataset contains three categories, cat, dog, and wildlife,  
and the evaluation is in the supplementary material.

\subsection{Attack Details}
\vspace{1mm}
\noindent\textbf{Fine-tuning Process and Evaluation.}
In Eq.~\eqref{eq:st_loss}, we set $\lambda$ to 1. We denote the number of iteration steps in a diffusion process, a fine-tuning process, and a sampling process as $s_{\textit{df}},s_{\textit{ft}}$, and $s_{\textit{sp}}$,
which are set to $40, 15$ and $40$ respectively.  For $s_{\textit{ft}}$, according to~\cite{kim2022diffusionclip}, even $6$ steps would satisfy the fine-tuning purpose. In our experiments, we set it to $15$ due to the VRAM limitation on our GPUs.  However, the semantic adversarial images would demonstrate a smoother modification with higher image quality by increasing the fine-tuning step. 
To ensure the adversary of our generated image, we initially verify the target classifier output throughout the fine-tuning process, and if the label remains unchanged, we will conduct additional $15$ steps of the fine-tuning process based on the last run iteratively until the attack is successful. The procedure of sampling will then be carried out to improve the image quality.

\vspace{1mm}
\noindent\textbf{Constructing Mask and Evaluation.}
Every pair of source and target images, $\mathbf{x}^s_0$ and 
\begin{wraptable}{r}{7cm}
\vspace{-3mm}
\centering
\adjustbox{width=0.55\textwidth}{
    \centering
    \begin{threeparttable}
    \noindent
    \begin{tabular}{l|c|ccccc}
    \toprule
    Setting & \begin{tabular}[c]{@{}l@{}}strategy\end{tabular}& ASR (\%)$\uparrow$ & FID$\downarrow$ & KID$\downarrow$ & \begin{tabular}[c]{@{}l@{}}average \\query$\downarrow$\end{tabular} & \begin{tabular}[c]{@{}l@{}}average\\ time (s)$\downarrow$\end{tabular} \\
    \midrule
    clean images & - & - & 30.67 & 0.000 & - & - \\ 
    \midrule
    LatentHSJA & - & 100.0 & 83.52 & 0.046 & 1000$^\dagger$ & 45.87 \\ 
    AttAttack & - & 71.80 & 48.92  & 0.018  & 146.82 & 49.71\\ 
    \midrule[1pt]
    \midrule[1pt]
   \multicolumn{7}{c}{ST approach} \\
    \midrule
    \multirow{2}{*}{\begin{tabular}[c]{@{}l@{}}fine-tune \\latent space\end{tabular}} 
    & white-box & \textbf{100.0} & 37.93 & 0.014 &7.72 & 37.10 \\ 
    & black-box & 59.18 & 114.99 & 0.098& 43.15 & 206.13 \\ 
    \midrule
    \multirow{2}{*}{\begin{tabular}[c]{@{}l@{}}fine-tune \\ diffusion  model\end{tabular}} & white-box & 99.2 & \textbf{36.61} & \textbf{0.006} & 4.98 & \textbf{30.78} \\ 
     & black-box & \textbf{100.0} & 96.88 & 0.068 & 11.73 & 66.57 \\ 
     \midrule
    \multirow{2}{*}{\begin{tabular}[c]{@{}l@{}}fine-tune both\end{tabular}} & white-box & 99.4 & 36.66 & \textbf{0.006} & \textbf{4.96} & \textbf{30.78} \\ 
     & black-box & \textbf{100.0} & 94.36 & 0.066 & 11.672 & 64.97 \\ 

    \midrule[1pt]
    \midrule[1pt]
       \multicolumn{7}{c}{LM approach} \\
    \midrule
    \multirow{3}{*}{GradCAM} 
    & $\mathbf{\hat{m}}_{s}(\delta)$ & 98.8 & 65.84 & 0.015 & 15.33 & 20.96 \\ 
     & $\mathbf{\hat{m}}_{t}(\delta)$ & 99.2 & \textbf{64.38} & \textbf{0.014} & 15.21 & \textbf{18.89} \\

    & $\mathbf{\hat{m}}_{s+t}(\delta)$ & 99.0 & 65.47 & \textbf{0.014} & \textbf{14.65} & 20.81\\ 
    \midrule
    \multirow{3}{*}{SimpleFullGrad} & $\mathbf{\hat{m}}_{s}(\delta)$ & 99.6 & 67.10 & 0.016 & 16.17 & 24.03 \\ 
     & $\mathbf{\hat{m}}_{t}(\delta)$ & 99.6 & 65.21 & 0.016 & 15.32 & 27.48\\ 
     & $\mathbf{\hat{m}}_{s+t}(\delta)$ & \textbf{99.8} & 65.67 & 0.015 & 14.73 & 23.77 \\ 
    \bottomrule
    \end{tabular}
    \begin{tablenotes}
    \item[$\dagger$]\, {Elapsed time varies, depending on the query steps, which is preset by the user.} \\
    \end{tablenotes}
    \end{threeparttable}
    }
    \vspace{-4mm}
    \caption{
    Performance of our framework with the ST and the LM approach on CelebA-HQ dataset compared with other two baselines.}
    \label{tab:framework}
\end{wraptable} 
$\mathbf{x}^t_0$, is randomly sampled as long as they have different class labels and can be classified correctly by the target classifier. 
When applying Grad-CAM or saliency map to an image, we combine $\mathbf{m}_0$ and $\mathbf{m}_1$ from the RGB channels into one channel and then filter by $\textit{TopK}$ in order to better observe how features are transplanted from the target image to the original clean image. For the saliency map, 
we adopt SimpleFullGrad from~\cite{srinivas2019full}. 
For the Grad-CAM~\cite{selvaraju2017grad}, we directly use the original implementation. 

\noindent After integrating the generated mask
 from Eq.~\eqref{eq:lm_mask_generation} and Eq.~\eqref{eq:lm_latent_generation}, the diffusion process does not always generate adversarial examples, and we need to decrement hyper-parameter $\delta$ in every iteration with Eq.~\eqref{eq:delta_decremental_speed}. 

In this approach, semantic adversarial 
images are generated using only the mask and a target image without the fine-tuning process. We set both $s_{\textit{df}}$ and $s_{\textit{sp}}$ to $40$. During each iteration, with decremental $\delta$, we check if the generated image from the sampling process is an adversarial example against the targeted classifier $f$, and we stop the attack when an adversarial example is generated.

\subsection{Evaluation Metrics and Benchmarks}
We quantify the attack success rate (ASR), Fréchet Inception Distance (FID)~\cite{heusel2017gans} and Kernel Inception Distance (KID)~\cite{binkowski2018demystifying} in the fidelity of semantic adversarial attacks. FID measures the Fréchet distance between two data distributions, and KID measures the dissimilarity between two distributions. For both measurements, lower is better.
In addition, for the ST method, we measure the average number of fine-tuning iterations, denoted as $\delta_{avg}$; for the LM method, we measure the average threshold for successful semantic adversarial attacks, denoted as $\eta_{avg}$. 
We also measure the average elapsed time for generating a semantic adversarial attack to evaluate efficiency.

\subsection{Results}
\label{sec:results}

\begin{wrapfigure}{r}{7cm}\vspace{-8mm}
\centering
\renewcommand*{\arraystretch}{1.0}
\begin{tabular}{>{\centering}m{0.043\textwidth}  >{}m{0.4\textwidth}}

& 
\hspace{14mm} \footnotesize{LM approach with Grad-CAM}
\\
{\small $\mathbf{\hat{m}}_{{s}}(\delta)$}
& 
\includegraphics[width=0.4\textwidth]{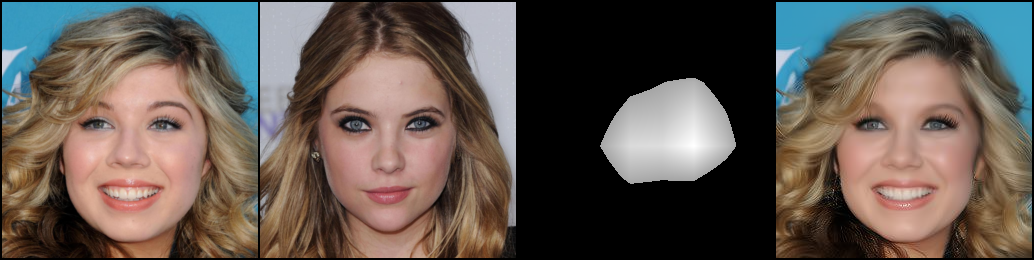}
\vspace{-5.1mm}
\\
{\small $\mathbf{\hat{m}}_{{t}}(\delta)$}
& 
\includegraphics[width=0.4\textwidth]{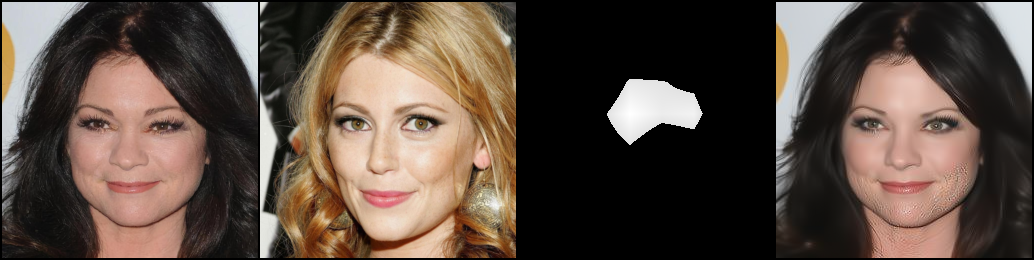}
\vspace{-5.1mm}
\\
{\footnotesize $\mathbf{\hat{m}}_{{s+t}}(\delta)$}
& 
\includegraphics[width=0.4\textwidth]{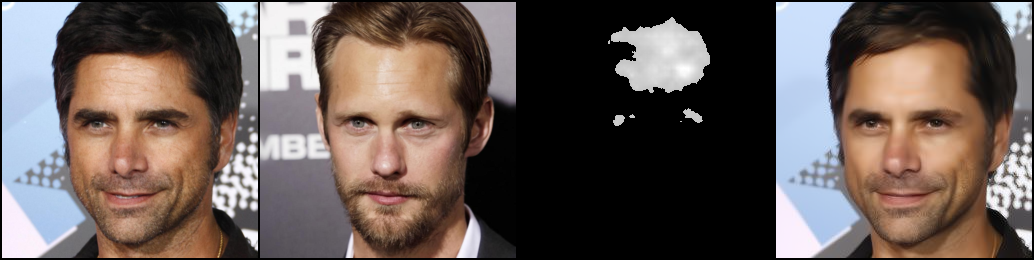}
\\
& 
\hspace{10mm} \footnotesize{LM approach with SimpleFullGrad}
\\
{\small $\mathbf{\hat{m}}_{{s}}(\delta)$}
& 
\includegraphics[width=0.4\textwidth]{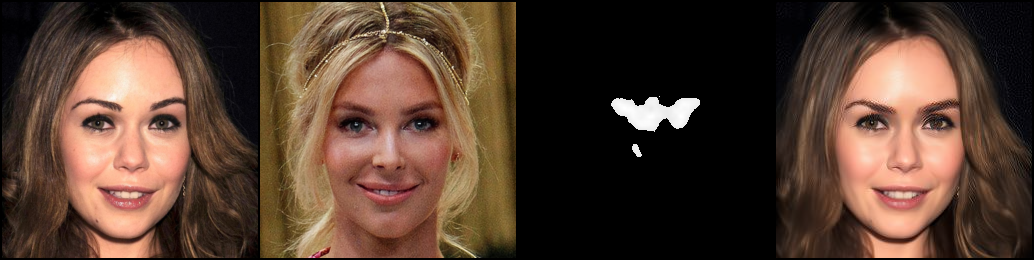}
\vspace{-5.1mm}
\\
{\small $\mathbf{\hat{m}}_{{t}}(\delta)$}
& 
\includegraphics[width=0.4\textwidth]{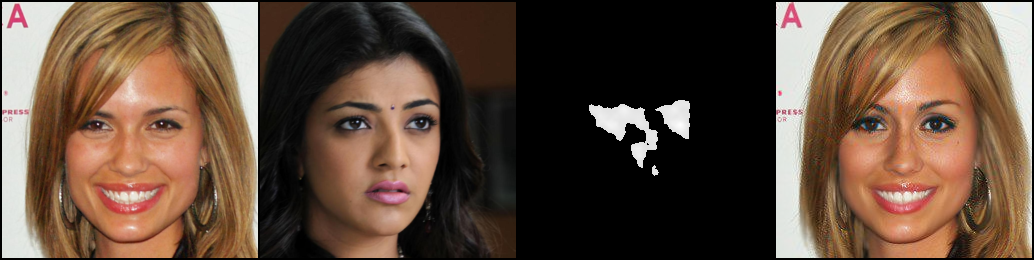}
\vspace{-5.1mm}
\\
{\footnotesize $\mathbf{\hat{m}}_{{s+t}}(\delta)$}
& 
\includegraphics[width=0.4\textwidth]{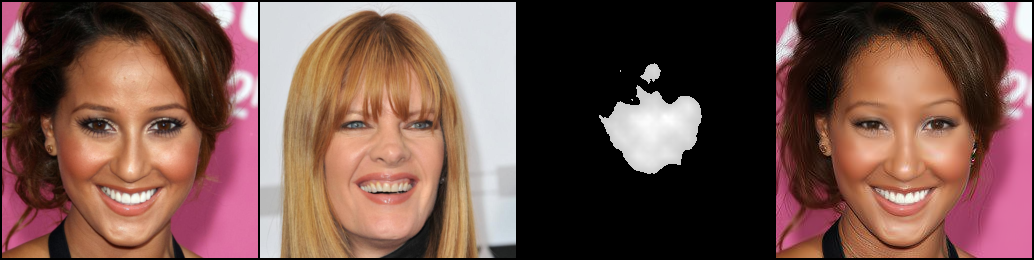}
\end{tabular}
\vspace{2mm}
\caption{
LM approach with 
three strategies 
$\mathbf{\hat{m}}_{s}(\delta)$, $\mathbf{\hat{m}}_{t}(\delta)$ and 
$\mathbf{\hat{m}}_{{s+t}}(\delta)$. From left to right: source images, target images, constructed masks with given strategies and generated semantic adversarial images. 
\label{fig:lm_1}
}
\vspace{-3mm}
\end{wrapfigure} 
The results and analysis focus on CelebA-HQ identity dataset, for which the results with the ST and the LM approach of our framework are shown in Table~\ref{tab:framework}. For baselines, we use two groups of clean images to calculate FID and KID, each group with 500 images. For comparison, we use recent semantic adversarial attacks LatentHSJA~\cite{na2022unrestricted} and AttAttack~\cite{joshi2019semantic} as benchmarks. For LatentHSJA, we run experiments with default fixed 1,000 queries. The number of queries is preset by 
 the user, and the default is 20,000. We use 1,000 as a trade-off between quality and efficiency. For AttAttack, we run several benchmarks of it and choose AttGAN to perturb the Age attribute, since this setting balances quality and ASR. In addition, we use the same target model in AttAttack as in our attacks. Both baselines are evaluated on Celeb-HQ Facial Identity Recognition Dataset as mentioned in Section~\ref{sec:datasets}. 
For all experiments, we focus on untargeted attacks, that is, as long as the predicted label is different from the original one, the attack will be counted as a success.

\vspace{1mm}
\noindent\textbf{Analysis for the ST Approach.}
Table~\ref{tab:framework} represents the performance of our framework on CelebA-HQ dataset. The top half shows the statistics of our framework with the ST approach under white-box and black-box settings. Our attack achieves almost 100\% ASR in all cases. For the ST approach, from FID and KID, we can clearly observe that our framework under white-box settings obtains higher-quality images than black-box settings. Under black-box settings, our framework creates more deformation than white-box settings and makes the generated images unrealistic and dissimilar to the original image. Visual examples generated with the ST approach are in Figure~\ref{fig:st}.

\begin{wrapfigure}{r}{0.5\textwidth}\vspace{-11mm}
\begin{center}
\includegraphics[width=0.5\textwidth]{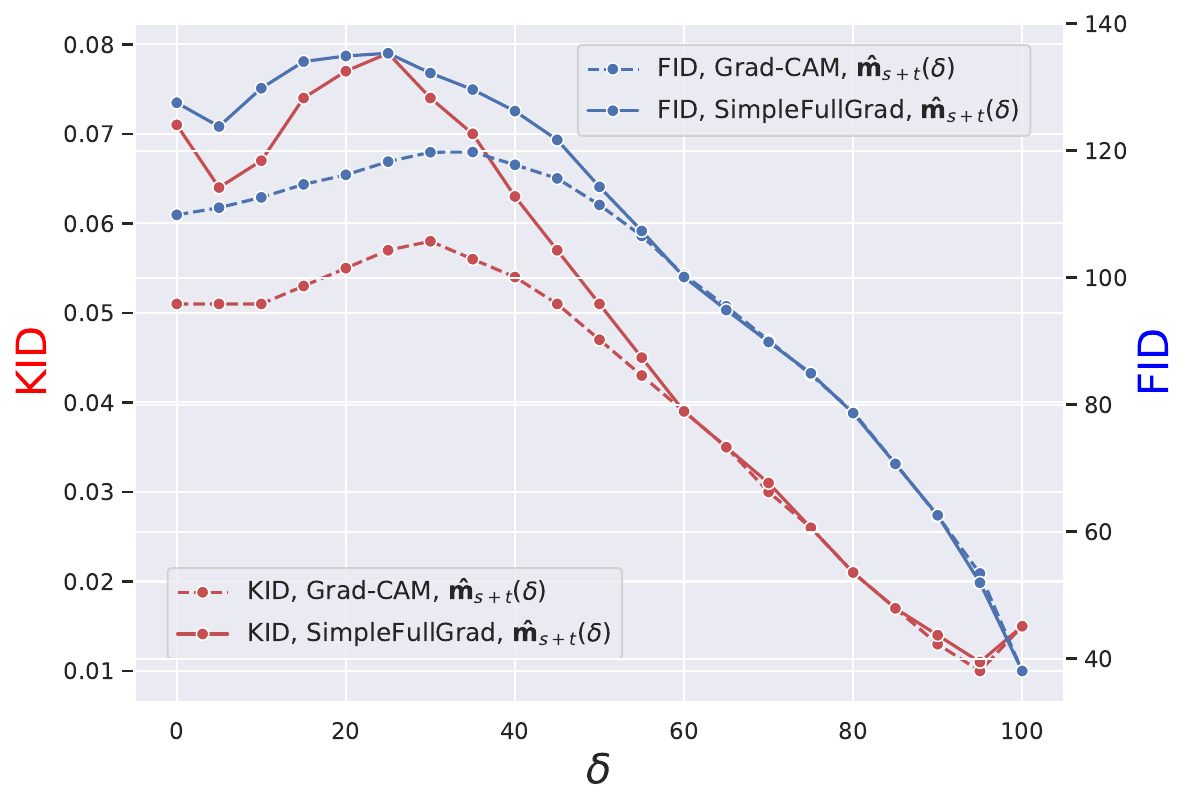}
      \vspace{-5mm}
  \caption{
  The relationship between FID/KID and $\delta$ in the LM approach of our framework. 
  }
  \label{fig:fid_kid_relationship}
  \vspace{0mm}
\end{center}
\end{wrapfigure}

From these generated images with the ST approach, we  find that our framework under white-box settings tends to achieve a minimal amount of modification concentrated in a small area of the original image. Under black-box settings, our framework tends to randomly modify details on a relatively large scale. 
\begin{wrapfigure}{r}{0.5\textwidth}\vspace{-11mm}
\begin{center}
\includegraphics[width=0.5\textwidth]{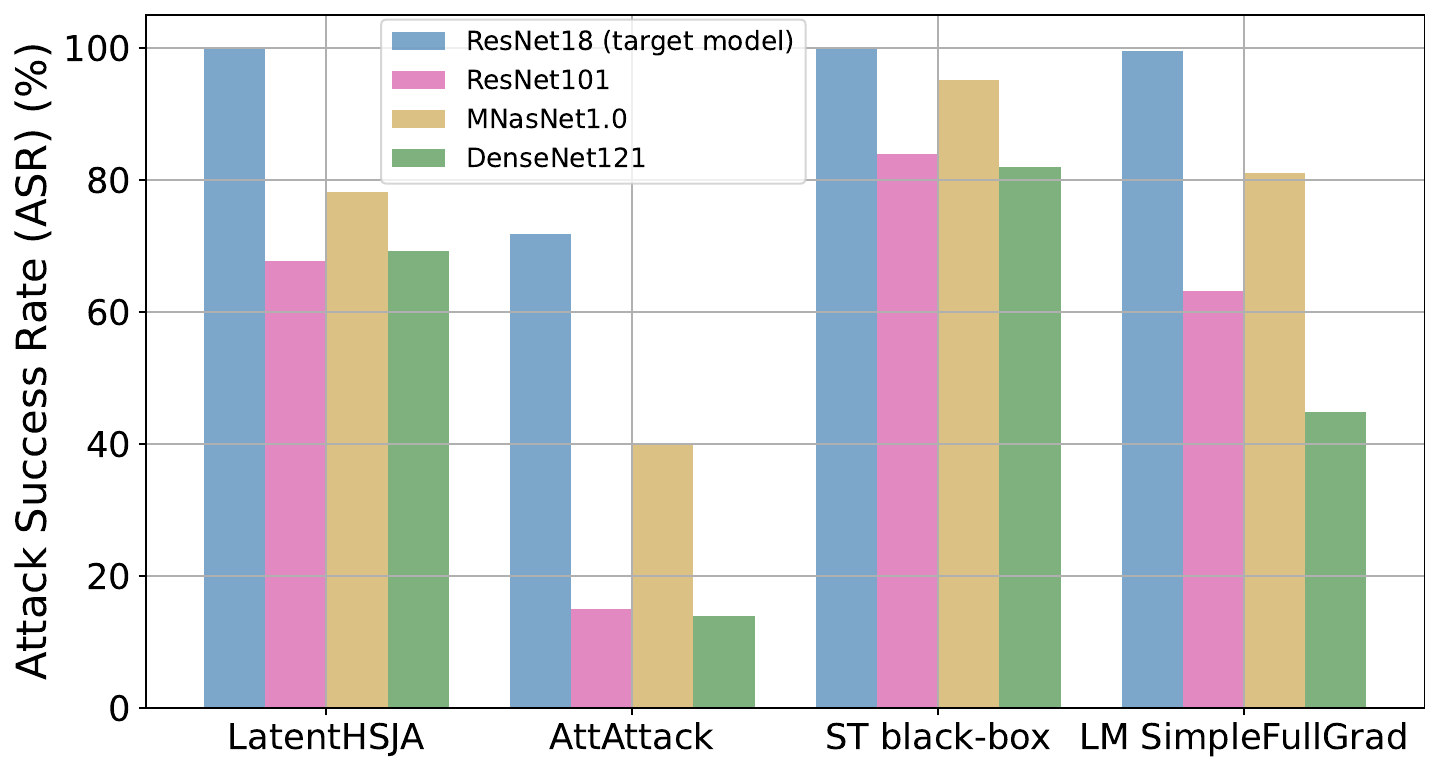}
      \vspace{-5mm}
  \caption{
  Transfer attack results on LatentHSJA, AttAttack, our ST and LM approach.}
  \label{fig:transfer_asr_1}
  \vspace{-7mm}
\end{center}
\end{wrapfigure}
This can be explained by the classifier $f$ in the KL divergence loss term $\mathit{D}_{\textit{KL}}$ of Eq.~\eqref{eq:st_loss}. Under white-box settings, $f$ is our targeted classifier in ResNet18, whereas under black-box settings, $f$ is a pretrained InceptionV3 network on ImageNet~\cite{deng2009imagenet}. Thus, $f$ could not efficiently capture the most important area with respect to a face identity under black-box settings. In addition, we find that fine-tuning both a latent space and a diffusion model achieves the best balance between quality and efficiency. 

\vspace{1mm}
\noindent\textbf{Analysis for the LM Approach.}
We found that constructing masks using GradCAM or SimpleFullGrad saliency maps yields similar results like nearly 100\% ASR and similar FIDs and KIDs. In most human faces, the features related to a face's identity usually come to be in a similar area on an image (e.g. nose, eyes, chin, and forehead). 
Hence, we can transplant features by directly applying masks to source and target images.

Visual examples are shown in Figure~\ref{fig:lm_1}.
Note that the mask may not exactly correspond to the area being modified since the generative process is performed via a diffusion model, the Gaussian noise added during the sampling process cannot precisely only modify a specified area without special techniques such as image inpainting. However, the most significant modifications occur in the mask areas overall. 
We also investigate the sensitivity of masking threshold $\delta$, as shown in Figure~\ref{fig:fid_kid_relationship}, and we find that the LM approach with Grad-CAM achieves better fidelity and quality than SimpleFullGrad in the same $\delta$, This implies that Grad-CAM has a better ability to expose attackable areas than SimpleFullGrad. 

\vspace{1mm}
\noindent\textbf{Comparison with Benchmarks.}
Visual examples for comparison are shown in Figure~\ref{fig:visuals_1}, and more of them can be found in 
the supplementary material. 
In terms of FID and KID, our framework with the ST approach under white-box settings achieves comparable performance. For average elapsed time per generated image, our framework with the LM approach shows the best performance compared to other methods. 
LatentHSJA starts
\begin{wrapfigure}{r}{6.5cm}
\vspace{-2mm}
\centering
\renewcommand*{\arraystretch}{0}
\begin{tabular}{*{5}{@{}c}@{}}
\includegraphics[scale=0.14]{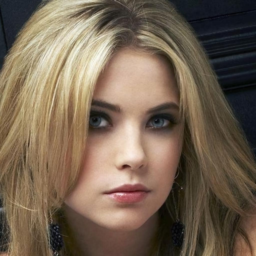}    & 
\includegraphics[scale=0.14]{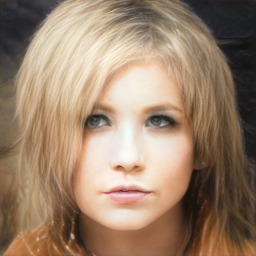}   & 
\includegraphics[scale=0.14]{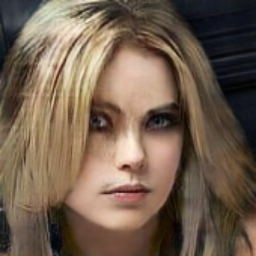} &
\includegraphics[scale=0.14]{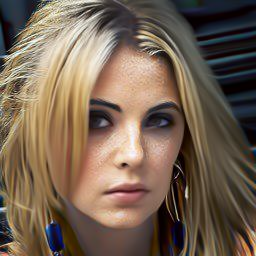}
  & 
\includegraphics[scale=0.14]
{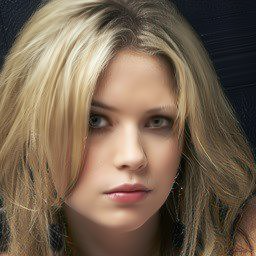} 
\\
\includegraphics[scale=0.14]{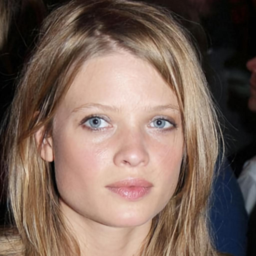}    & 
\includegraphics[scale=0.14]{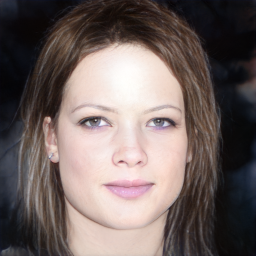}   & 
\includegraphics[scale=0.14]{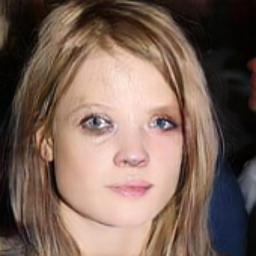} &
\includegraphics[scale=0.14]{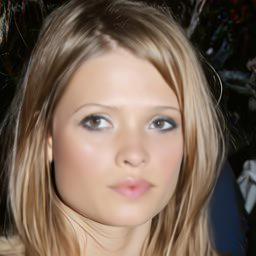}  
  & 
\includegraphics[scale=0.14]{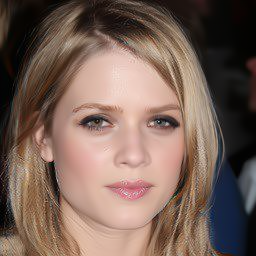} 
\\
\includegraphics[scale=0.14]{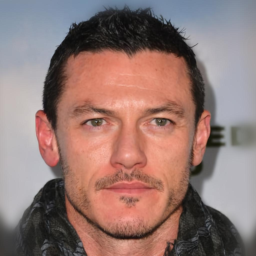}    & 
\includegraphics[scale=0.14]{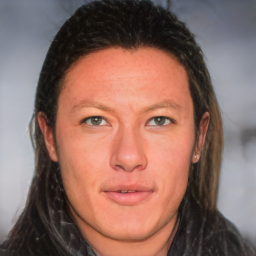}   & 
\includegraphics[scale=0.14]{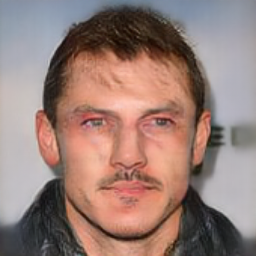} &
\includegraphics[scale=0.14]{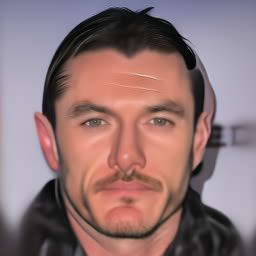}  
  &
\includegraphics[scale=0.14]{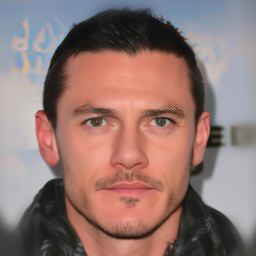} 
\\
\includegraphics[scale=0.14]{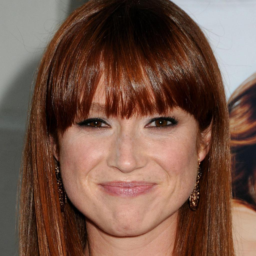}    & 
\includegraphics[scale=0.14]{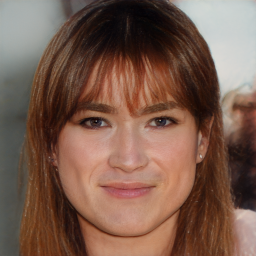}   & 
\includegraphics[scale=0.14]{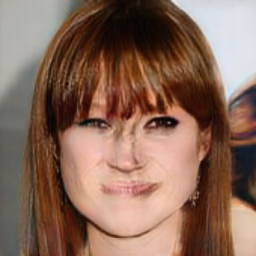} &
\includegraphics[scale=0.14]{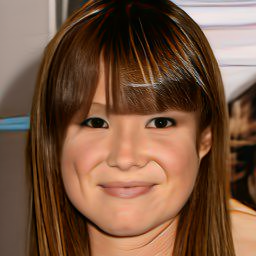}  
  &
\includegraphics[scale=0.14]{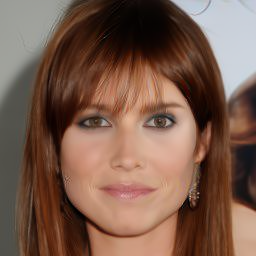} 
\\
\includegraphics[scale=0.14]{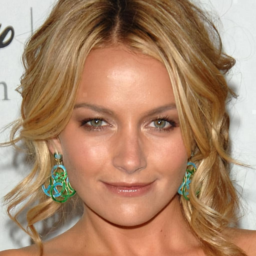}    & 
\includegraphics[scale=0.14]{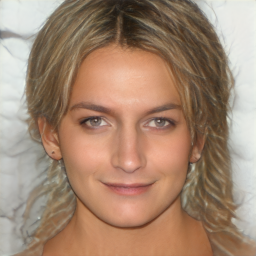}   & 
\includegraphics[scale=0.14]{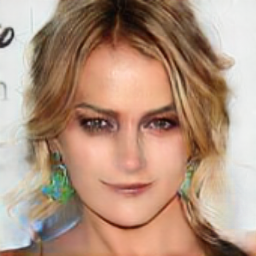} &
\includegraphics[scale=0.14]{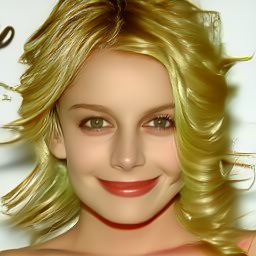}  
  &
\includegraphics[scale=0.14]{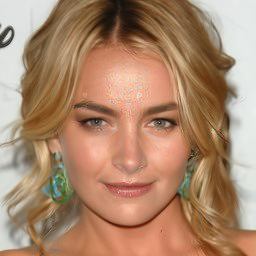} 
\\
\includegraphics[scale=0.14]{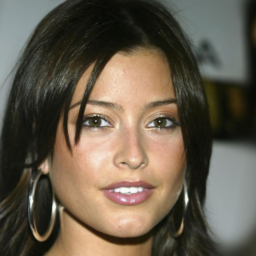}    & 
\includegraphics[scale=0.14]{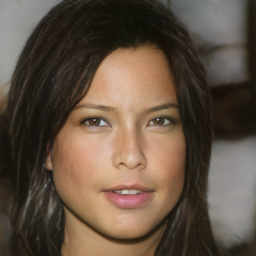}   & 
\includegraphics[scale=0.14]{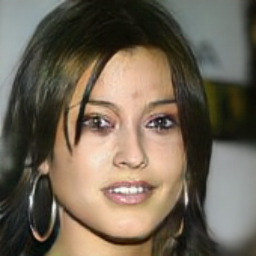} &
\includegraphics[scale=0.14]{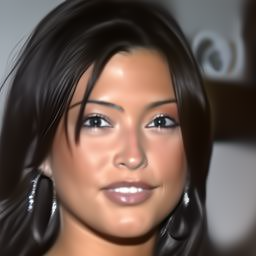}  
  &
\includegraphics[scale=0.14]{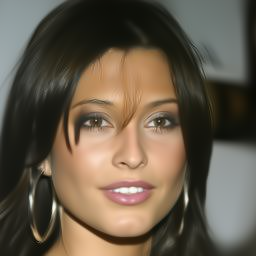} 
\end{tabular}
\vspace{3mm}
\caption{
Comparison of semantic adversarial attacks between ours and others.
From left to right:
original image, LatentHSJA, AttAttack, our ST black-box approach, and LM with Grad-CAM approach.}
\label{fig:visuals_1}
\vspace{-4mm}
\end{wrapfigure}
from the target image and iteratively tries to transplant pixels in latent space, with a fixed number of queries, but with a low similarity compared with the original image. AttAttack greatly improves performance by utilizing manual attribute annotations, yet it is challenging to generalize to other datasets. In addition, we evaluate the transferability in Figure~\ref{fig:transfer_asr_1}. 
All methods targeted their attack on Resnet18 (except for ST Black-box) and generated 500 adversarial images. We test these on three additional classifiers, ResNet101, MNasNet1.0, and DenseNet121 to calculate the ASR. Our ST black-box approach performs the best among the others as expected, since it does not require exact information from the target model. Compared with the ST approach, the generated adversarial attacks should be harder to denoise by diffusion-based purification algorithms such as \cite{nie2022diffusion,wang2022guided,wu2022guided}. 
We further evaluate the robustness of our semantic attack under natural perturbations, such as JPEG, Gaussian Blur, Defocus Blur, and Brightness transformation in the supplementary material. 
The results demonstrate that our semantic adversarial perturbations are still preserved after natural perturbations.

\vspace{3mm}
\section{Conclusion}
In this paper, we first proposed a framework for semantic adversarial attacks by leveraging Diffusion Models with the ST approach and LM approach. The ST approach manipulates the latent space of a benign image or the parameters of a diffusion model via fine-tuning, whereas the LM approach manipulates the latent space via masking of significance maps in a more direct way. In our empirical study, the proposed framework achieves excellent performance for both approaches, under different settings. In total, our framework shows great generalizability, efficiency, and transferability compared to other baselines and exposes a novel usage of Diffusion Models in the semantic adversarial attack domain.
However, there are limitations to our framework. For the ST approach, the quality of adversarial images under the black-box setting is not as good as the white-box setting; for the LM approach, it cannot precisely control the masked area to be modified compared to a clean image. 
\clearpage
\bibliography{egbib}
\include{Appendix_final.tex}
\end{document}

%% file: Appendix_final.tex
\appendix
\section*{Supplementary Material}
\section{Algorithms of our framework}\label{sec:algos}
The ST and LM approaches of our framework are shown in Algorithm~\ref{alg:st} and Algorithm~\ref{alg:lm}, respectively. 

\begin{algorithm}[H]
\caption{ST approach}%
\label{alg:st}
\begin{algorithmic}[1]
  \REQUIRE{Input image $\mathbf{x}_0$, diffusion model with weight parameters $\theta$, number of iterations $N$, $s_{\textit{ft}}$ and $s_{\textit{sp}}$;}
  \ENSURE{Semantic adversarial image $\mathbf{\hat{x}}_0$;}
  \STATE{$\mathbf{x}_T \leftarrow$ update by Eq.~\eqref{eq:generated_latent} with $\mathbf{x}_0$;} 
  \STATE{Initialize $\mathbf{\hat{x}}_{T} \leftarrow \mathbf{x}_T$, $\hat{\theta} \leftarrow \theta$;}
  \FOR{$i =0 $ \TO $N - 1$}
  \STATE{$\mathbf{\hat{x}}_0\leftarrow$ update by Eq.~\eqref{eq:generated_image} with $\hat{\theta}$ and $\mathbf{\hat{x}}_T$ in $s_{\textit{ft}}$ steps;}
  \STATE{Optimize $\mathbf{\hat{x}}_{T}$ and/or $\hat{\theta}$ by Eq.~\eqref{eq:st_loss};}
    \IF{$f(\mathbf{\hat{x}}_0) \neq f(\mathbf{x}_0)$}
    \STATE{$\mathbf{\hat{x}}_0\leftarrow$ update by Eq.~\eqref{eq:generated_image} with $\hat{\theta}$ and $\mathbf{\hat{x}}_T$ in $s_{\textit{sp}}$ steps;}
    \IF{$f(\mathbf{\hat{x}}_0) \neq f(\mathbf{x}_0)$}
    \STATE{Early stop and return $\mathbf{\hat{x}}_0$;}
    \ENDIF 
  \ENDIF
  \ENDFOR
  \STATE{$\mathbf{\hat{x}}_0\leftarrow$ update by Eq.~\eqref{eq:generated_image} with $\hat{\theta}$ and $\mathbf{\hat{x}}_T$ in $s_{\textit{sp}}$ steps;}
  \RETURN{$\mathbf{\hat{x}}_0$}
\end{algorithmic}
\vspace{-1mm}
\end{algorithm}
\begin{algorithm}[H]
\caption{LM approach}%
\label{alg:lm}
\begin{algorithmic}[1]
  \REQUIRE{A pair of source and target image as $\mathbf{x}^s_0$ and $\mathbf{x}^t_0$, an interpretation map function $g$, threshold $\delta$;}
  \ENSURE{Semantic adversarial image $\mathbf{\hat{x}}_0$;}
  \STATE{$\mathbf{x}^s_T, \mathbf{x}^t_T\leftarrow$ Eq.~\eqref{eq:generated_latent} with $\mathbf{x}^s_0$ and $\mathbf{x}^t_0$ respectively;}
  \STATE{Obtain original interpretation maps $\mathbf{m}_s$ and $\mathbf{m}_t$ with function $\mathbf{m} = g(\mathbf{x}_0,\mathbf{y})$;} 
  \STATE{Initialize $\delta \leftarrow 100$;}
  \WHILE{$\delta\neq0$}
  \STATE{$\mathbf{\hat{m}}\leftarrow$ update by  Eq.~\eqref{eq:lm_mask_generation} with $\mathbf{m}_s$ and $\mathbf{m}_t$;}
  \STATE{$\mathbf{\hat{x}}_{T}\leftarrow$ update by Eq.~\eqref{eq:lm_latent_generation} with $\mathbf{\hat{m}}$;}
  \STATE{$\mathbf{\hat{x}}_{T} \leftarrow$ update by Eq.~\eqref{eq:generated_image} with $\mathbf{\hat{x}}_0$;}
    \IF{$f(\mathbf{\hat{x}}_0) \neq f(\mathbf{x}_0)$}
    \STATE{Early stop and return $\mathbf{\hat{x}}_0$;}
    \ENDIF 
  \STATE{$\delta \leftarrow$ update by Eq.~\eqref{eq:delta_decremental_speed};}

  \ENDWHILE
  \STATE{$\mathbf{\hat{x}}_0 \leftarrow$ update by Eq.~\eqref{eq:generated_image}  with $\mathbf{\hat{x}}_{T}$;}
  \RETURN{$\mathbf{\hat{x}}_0$}
\end{algorithmic}
\end{algorithm}

\section{Additional Evaluations}\label{sec:additional results}

\subsection{Evaluation on Facial Identity Recognition}\label{sec:exp_id_more}

\begin{figure*}[]
\centering
\renewcommand*{\arraystretch}{0}
\begin{tabular}{*{9}{@{}c}@{}}
{Original} & \thead{LatentHSJA} & \thead{AttAttack} &\thead{ST\\black-box\\fine-tune\\model} & \thead{ST\\black-box\\fine-tune\\both}  & \thead{ST\\white-box\\fine-tune\\latent\\space}  & \thead{ST\\white-box\\fine-tune\\both} & \thead{LM\\Grad-CAM\\$\mathbf{\hat{m}}_{{s+t}}(\delta)$} & \thead{LM\\\tiny{SimpleFullGrad}\\$\mathbf{\hat{m}}_{{s+t}}(\delta)$} \\
\includegraphics[width=0.11\textwidth,keepaspectratio]{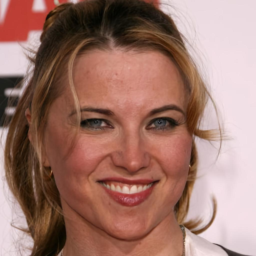}&
\includegraphics[width=0.11\textwidth,keepaspectratio]{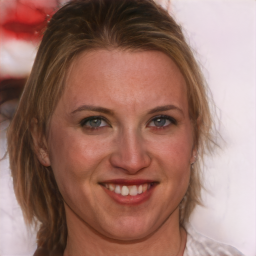}&
\includegraphics[width=0.11\textwidth,keepaspectratio]{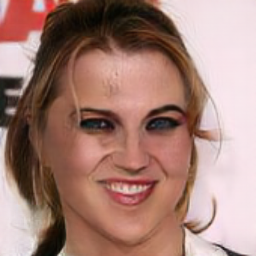}&
\includegraphics[width=0.11\textwidth,keepaspectratio]{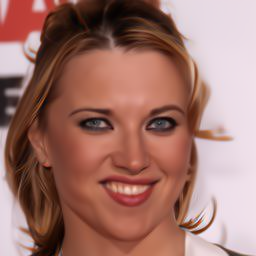}&
\includegraphics[width=0.11\textwidth,keepaspectratio]{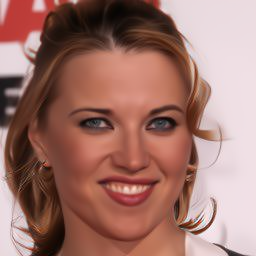}&
\includegraphics[width=0.11\textwidth,keepaspectratio]{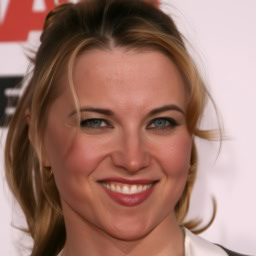}&
\includegraphics[width=0.11\textwidth,keepaspectratio]{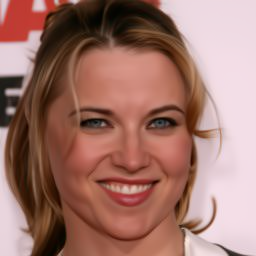}&
\includegraphics[width=0.11\textwidth,keepaspectratio]{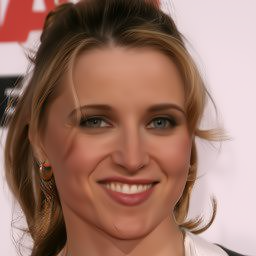}&
\includegraphics[width=0.11\textwidth,keepaspectratio]{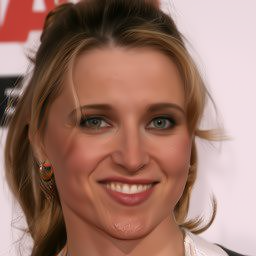}\\
\includegraphics[width=0.11\textwidth,keepaspectratio]{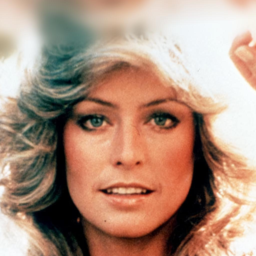}&
\includegraphics[width=0.11\textwidth,keepaspectratio]{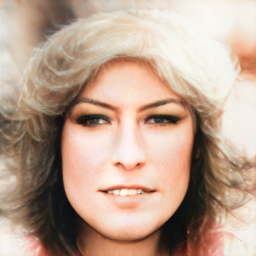}&
\includegraphics[width=0.11\textwidth,keepaspectratio]{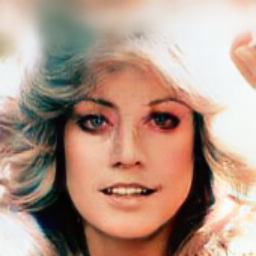}&
\includegraphics[width=0.11\textwidth,keepaspectratio]{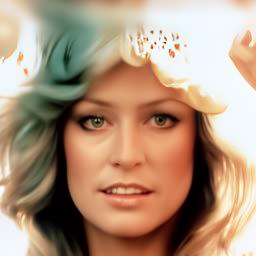}&
\includegraphics[width=0.11\textwidth,keepaspectratio]{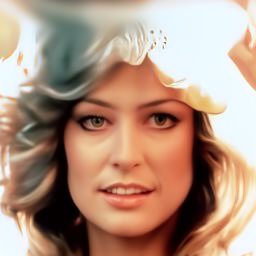}&
\includegraphics[width=0.11\textwidth,keepaspectratio]{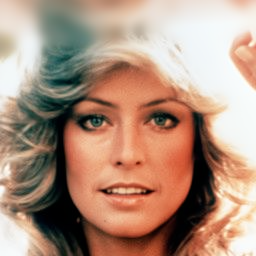}&
\includegraphics[width=0.11\textwidth,keepaspectratio]{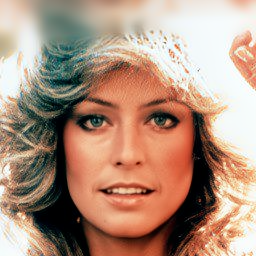}&
\includegraphics[width=0.11\textwidth,keepaspectratio]{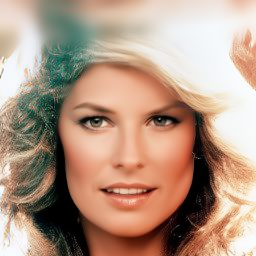}&
\includegraphics[width=0.11\textwidth,keepaspectratio]{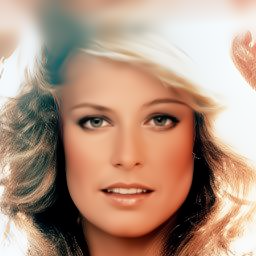}\\
\includegraphics[width=0.11\textwidth,keepaspectratio]{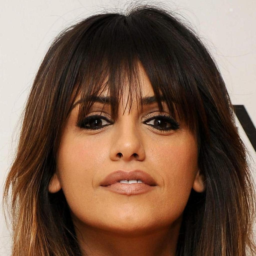}&
\includegraphics[width=0.11\textwidth,keepaspectratio]{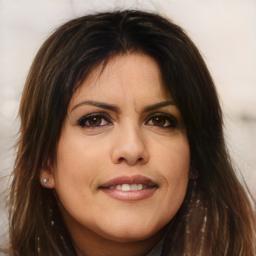}&
\includegraphics[width=0.11\textwidth,keepaspectratio]{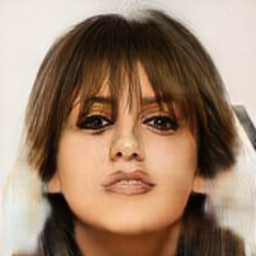}&
\includegraphics[width=0.11\textwidth,keepaspectratio]{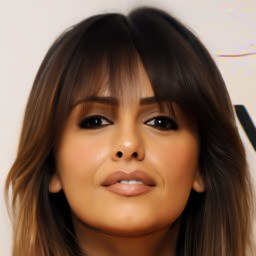}&
\includegraphics[width=0.11\textwidth,keepaspectratio]{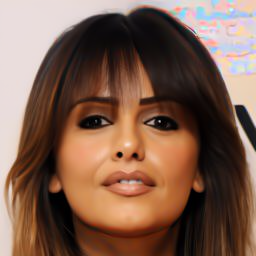}&
\includegraphics[width=0.11\textwidth,keepaspectratio]{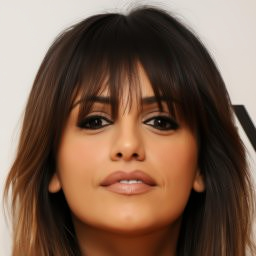}&
\includegraphics[width=0.11\textwidth,keepaspectratio]{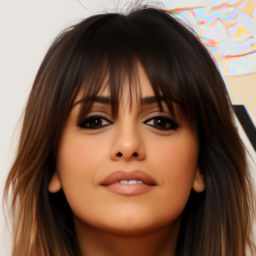}&
\includegraphics[width=0.11\textwidth,keepaspectratio]{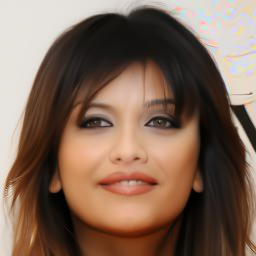}&
\includegraphics[width=0.11\textwidth,keepaspectratio]{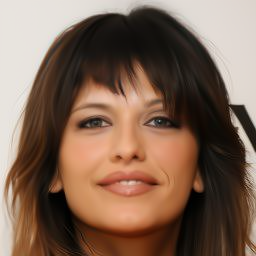}\\
\includegraphics[width=0.11\textwidth,keepaspectratio]{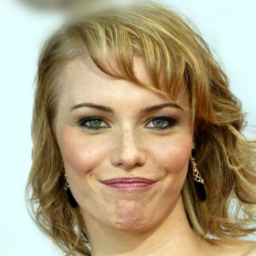}&
\includegraphics[width=0.11\textwidth,keepaspectratio]{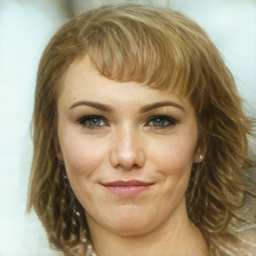}&
\includegraphics[width=0.11\textwidth,keepaspectratio]{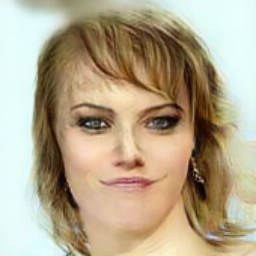}&
\includegraphics[width=0.11\textwidth,keepaspectratio]{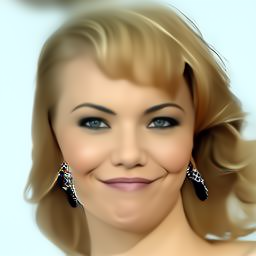}&
\includegraphics[width=0.11\textwidth,keepaspectratio]{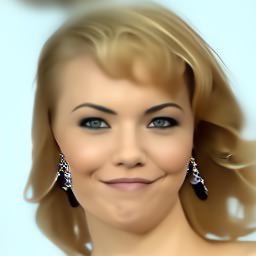}&
\includegraphics[width=0.11\textwidth,keepaspectratio]{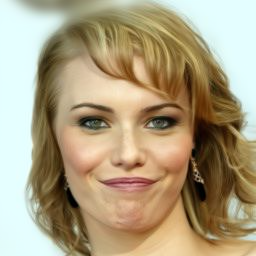}&
\includegraphics[width=0.11\textwidth,keepaspectratio]{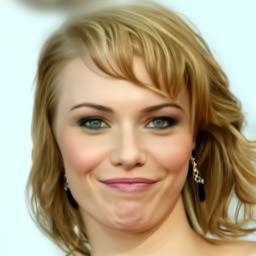}&
\includegraphics[width=0.11\textwidth,keepaspectratio]{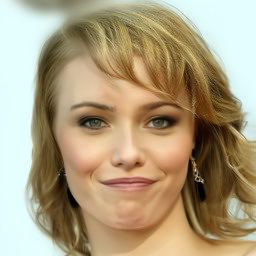}&
\includegraphics[width=0.11\textwidth,keepaspectratio]{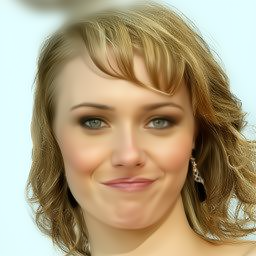}
\end{tabular}
\vspace{3mm}
\caption{Comparison of generated images between different methods on Celeb-HQ facial identity recognition dataset. For AttAttack, the perturbed attribute is Age.}
\label{fig:app_a} 
\end{figure*}

All experiments are executed on a single NVIDIA A40 GPU. The classifier is ResNet-18 with 89.63\% accuracy. For AttAttack~\cite{joshi2019semantic} on Celeb-HQ facial identity recognition dataset~\cite{na2022unrestricted} in this paper, the perturbed attribute is Age to balance quality and ASR. For LatentHSJA~\cite{na2022unrestricted}, we set the number of queries to 1,000 for the same reason. 
Visual examples for comparison between our methods and other methods are shown in 
Figure~\ref{fig:app_a}.
Our semantic attack accuracy before and after natural perturbations are shown in Table~\ref{table:ASR_robust}, where our semantic perturbations are still preserved against natural perturbations: even with the clean accuracy changes 47.8\% under Gaussian blur, our semantic attack accuracy only changes 14.6\%. 

\begin{table}[h!]
\centering
\adjustbox{width=0.8\textwidth}{
\begin{tabular}{c | c c c c c } 
 \toprule
 Setting & None& JPEG & Gaussian Blur & Defocus Blur & Brightness\\ 
 \midrule
 Clean & 100.0 & 100.0 & 52.2 & 95.4 & 100.0 \\
 \midrule
\begin{tabular}[c]{@{}l@{}}ST black-box\\fine-tune both\end{tabular} & 0.0 & 32.8 & 14.6 & 21.8 & 5.8 \\
\bottomrule
\end{tabular}
}
\vspace{3mm}
\caption{The accuracy for clean images and the semantic adversarial images (the ST approach, black-box, fine-tuning both the latent space and diffusion model) before and after natural perturbations. Most semantic adversarial perturbations are preserved.}
\label{table:ASR_robust}
\vspace{-3mm}
\end{table}

\subsection{Evaluation on Face Gender Classification}\label{sec:exp_gender}
\begin{figure*}[]
\centering
\renewcommand*{\arraystretch}{0}
\begin{tabular}{*{9}{@{}c}@{}}
{Original} & \thead{LatentHSJA} & \thead{AttAttack}  & \thead{LM\\Grad-CAM\\$\mathbf{\hat{m}}_{{s}}(\delta)$} &\thead{LM\\Grad-CAM\\$\mathbf{\hat{m}}_{{s+t}}(\delta)$} & \thead{LM\\\tiny{SimpleFullGrad} \\$\mathbf{\hat{m}}_{{s}}(\delta)$}& \thead{LM\\\tiny{SimpleFullGrad} \\$\mathbf{\hat{m}}_{{s+t}}(\delta)$}&\thead{ST\\white-box\\fine-tune\\latent\\space} & \thead{ST\\white-box\\fine-tune\\both} \\
\includegraphics[width=0.11\textwidth,keepaspectratio]{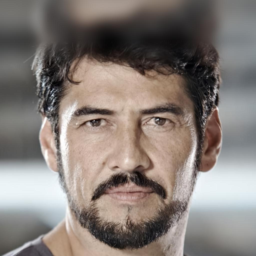}&
\includegraphics[width=0.11\textwidth,keepaspectratio]{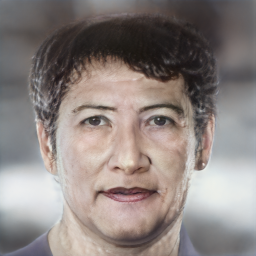}&
\includegraphics[width=0.11\textwidth,keepaspectratio]{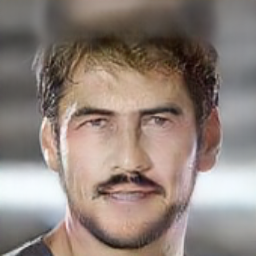}&
\includegraphics[width=0.11\textwidth,keepaspectratio]{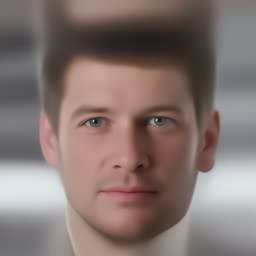}&
\includegraphics[width=0.11\textwidth,keepaspectratio]{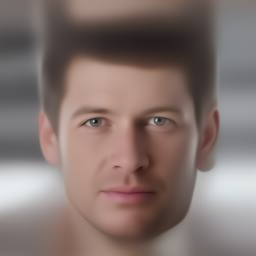}&
\includegraphics[width=0.11\textwidth,keepaspectratio]{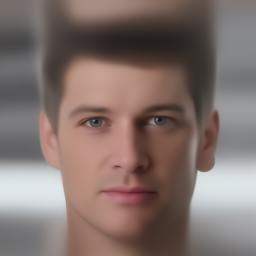}&
\includegraphics[width=0.11\textwidth,keepaspectratio]{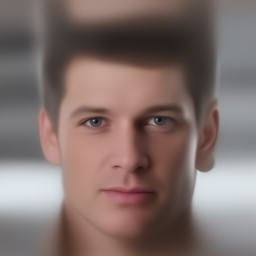}&
\includegraphics[width=0.11\textwidth,keepaspectratio]{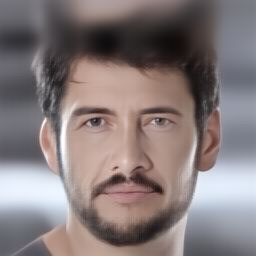}&
\includegraphics[width=0.11\textwidth,keepaspectratio]{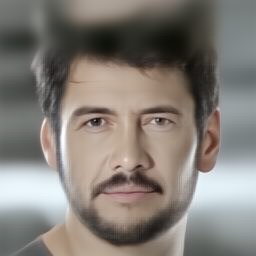}\\
\includegraphics[width=0.11\textwidth,keepaspectratio]{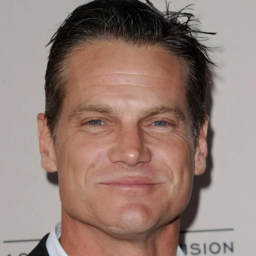}&
\includegraphics[width=0.11\textwidth,keepaspectratio]{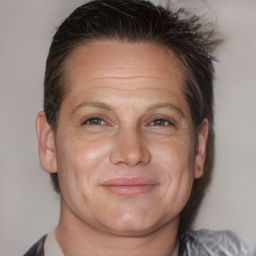}&
\includegraphics[width=0.11\textwidth,keepaspectratio]{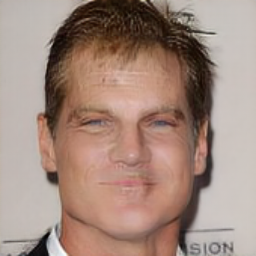}&
\includegraphics[width=0.11\textwidth,keepaspectratio]{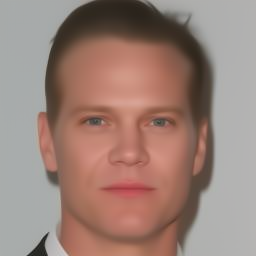}&
\includegraphics[width=0.11\textwidth,keepaspectratio]{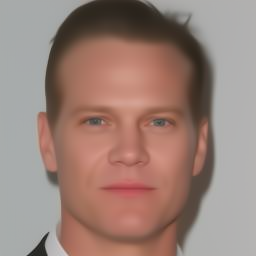}&
\includegraphics[width=0.11\textwidth,keepaspectratio]{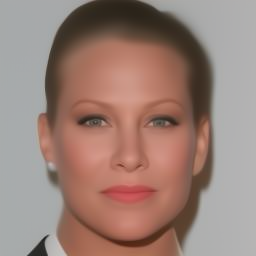}&
\includegraphics[width=0.11\textwidth,keepaspectratio]{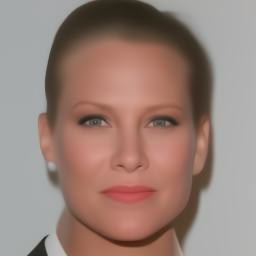}&
\includegraphics[width=0.11\textwidth,keepaspectratio]{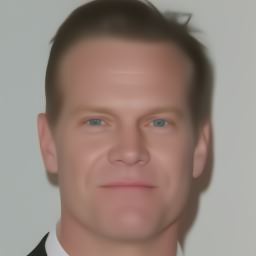}&
\includegraphics[width=0.11\textwidth,keepaspectratio]{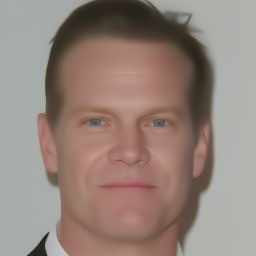}\\
\includegraphics[width=0.11\textwidth,keepaspectratio]{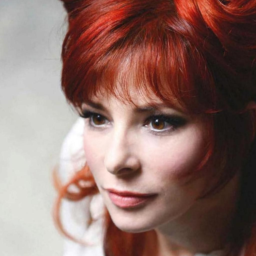}&
\includegraphics[width=0.11\textwidth,keepaspectratio]{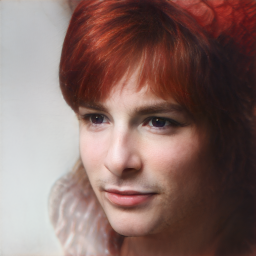}&
\includegraphics[width=0.11\textwidth,keepaspectratio]{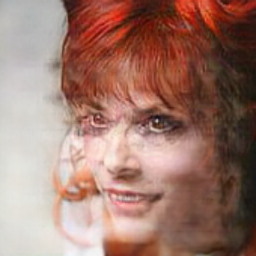}&
\includegraphics[width=0.11\textwidth,keepaspectratio]{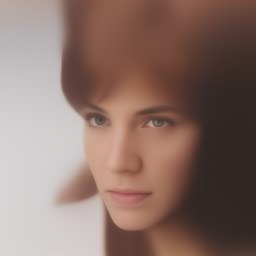}&
\includegraphics[width=0.11\textwidth,keepaspectratio]{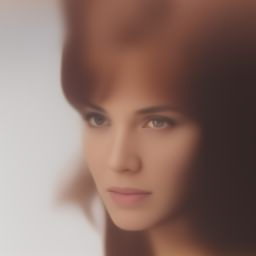}&
\includegraphics[width=0.11\textwidth,keepaspectratio]{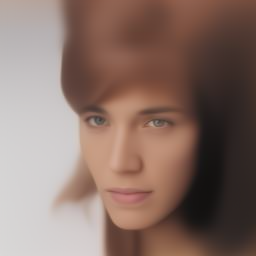}&
\includegraphics[width=0.11\textwidth,keepaspectratio]{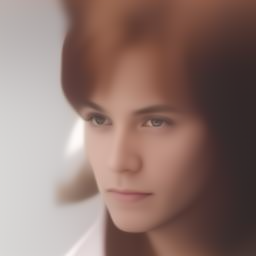}&
\includegraphics[width=0.11\textwidth,keepaspectratio]{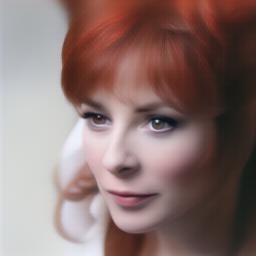}&
\includegraphics[width=0.11\textwidth,keepaspectratio]{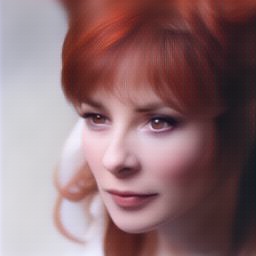}\\
\includegraphics[width=0.11\textwidth,keepaspectratio]{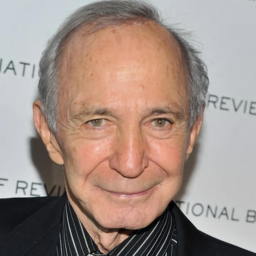}&
\includegraphics[width=0.11\textwidth,keepaspectratio]{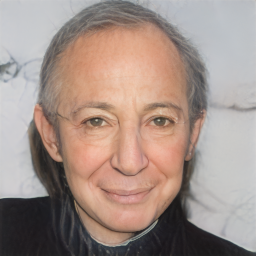}&
\includegraphics[width=0.11\textwidth,keepaspectratio]{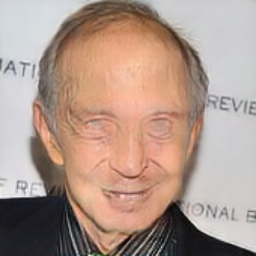}&
\includegraphics[width=0.11\textwidth,keepaspectratio]{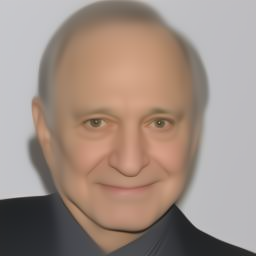}&
\includegraphics[width=0.11\textwidth,keepaspectratio]{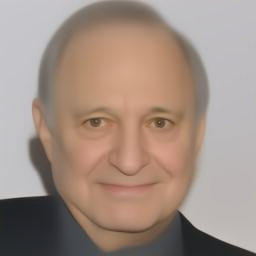}&
\includegraphics[width=0.11\textwidth,keepaspectratio]{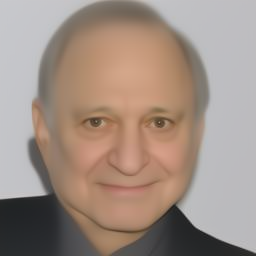}&
\includegraphics[width=0.11\textwidth,keepaspectratio]{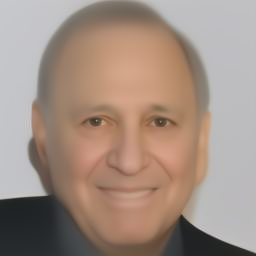}&
\includegraphics[width=0.11\textwidth,keepaspectratio]{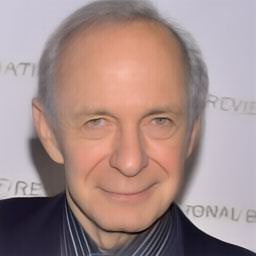}&
\includegraphics[width=0.11\textwidth,keepaspectratio]{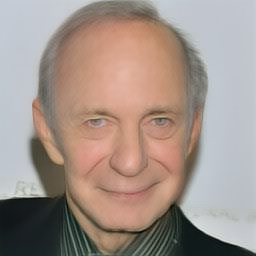}
\end{tabular}
\vspace{3mm}
\caption{Comparison of generated images between different methods on Celeb-HQ face gender classification dataset. For AttAttack attack, the perturbed attribute is Smiling.}
\label{fig:app_b}
\end{figure*}

The classifier is ResNet-18 with 98.38\% accuracy. For AttAttack~\cite{joshi2019semantic} on Celeb-HQ facial gender recognition dataset~\cite{na2022unrestricted}, the perturbed attribute is Smiling to balance quality and ASR. For LatentHSJA~\cite{na2022unrestricted}, we set the number of queries to 1,000 for the same reason. Results on Celeb-HQ facial gender recognition dataset with the ST approach are shown in Table~\ref{tab:app_b}, and visual examples for comparison between our framework and other methods are shown in Figure~\ref{fig:app_b}. 
\begin{table}[h]
\centering
\adjustbox{width=0.8\textwidth}{
    \centering
    \begin{threeparttable}
    \noindent
    \begin{tabular}{l|c|ccccc}
    \toprule
    Setting & \begin{tabular}[c]{@{}l@{}}strategy\end{tabular}& ASR (\%)$\uparrow$ & FID$\downarrow$ & KID$\downarrow$ & \begin{tabular}[c]{@{}l@{}}average \\query$\downarrow$\end{tabular} & \begin{tabular}[c]{@{}l@{}}average\\ time (s)$\downarrow$\end{tabular} \\
    \midrule
    clean images & - & - & 30.67 & 0.000 & - & - \\ 
    \midrule
    LatentHSJA & - & 100.0 & 28.64 & 0.005 & 1000 $^\dagger$ & 45.83  \\ 
    AttAttack, age & - & 34.2 & 72.99 & 0.029 & 373.20  & 94.43 \\ 
    AttAttack, 6 attrs & - &  85.80 & 96.31  & 0.060  & 78.86  & 26.14 \\ 
    \midrule[1pt]
    \midrule[1pt]
   \multicolumn{7}{c}{ST approach} \\
    \midrule
    \begin{tabular}[c]{@{}l@{}}fine-tune \\latent space\end{tabular} & white-box & 76.3 & 115.58 & 0.051 & 30.58 & 168.31 \\ 
    \midrule
    \multirow{2}{*}{\begin{tabular}[c]{@{}l@{}}fine-tune \\ diffusion  model\end{tabular}} 
    & white-box & 78.2  & 102.29 & 0.024 & 16.17 & \textbf{90.88} \\ 
     & black-box & 96.8 & 226.81 & 0.116 & 26.01 & 137.49 \\ 
     \midrule
    \multirow{2}{*}{\begin{tabular}[c]{@{}l@{}}fine-tune both\end{tabular}} 
    & white-box &    78.2   &    \textbf{98.19}    & \textbf{0.023} & \textbf{16.13}&90.96 \\ 
     & black-box & \textbf{97.4} & 235.45 & 0.119 & 26.02 & 137.59 \\ 

    \bottomrule
    \end{tabular}
    \begin{tablenotes}
    \item[$\dagger$]\, {Elapsed time varies, depending on the query steps, which is preset by the user.} \\
    \end{tablenotes}
    \end{threeparttable}
    }
    \caption{Our framework with the ST approach on CelebA-HQ gender classification dataset. For AttAttack attack, we run the AttAttack either on age attribute, or 6 attributes: Eyeglasses, Age, Pale Skin, Mustache, Eyebrows, and Hair Color.}
    \vspace{-2mm}
    \label{tab:app_b}
\end{table}

\subsection{Evaluation on AFHQ dataset}\label{sec:exp_afhq}
\begin{figure}[!htb]
\centering
\renewcommand*{\arraystretch}{0}
\begin{tabular}{*{5}{@{}c}@{}}
original & \thead{ST\\black-box\\fine-tune\\latent\\space}  &\thead{ST\\white-box\\fine-tune\\latent\\space}  &\thead{ST\\white-box\\fine-tune\\model}  &\thead{ST\\white-box\\fine-tune\\both}\\
\includegraphics[scale=0.19]{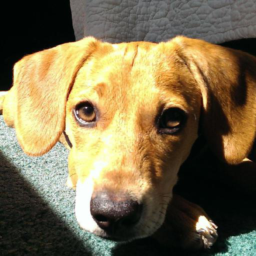}&
\includegraphics[scale=0.19]{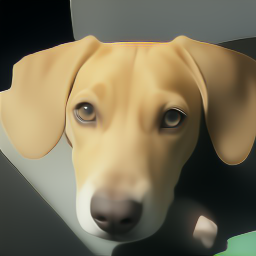}&
\includegraphics[scale=0.19]{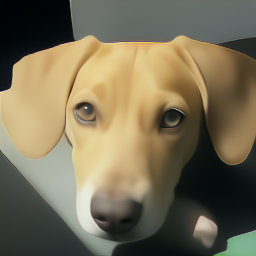}&
\includegraphics[scale=0.19]{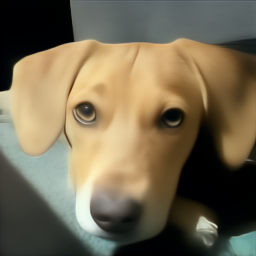}&
\includegraphics[scale=0.19]{2d/2_white_0_box_white_tune_0_mask9_diff9_gender_rec_train_2_100_only_reconstruction.png}
\\
\includegraphics[scale=0.19]{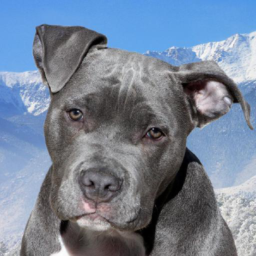}&
\includegraphics[scale=0.19]{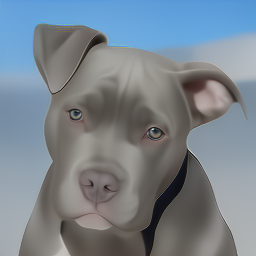}&
\includegraphics[scale=0.19]{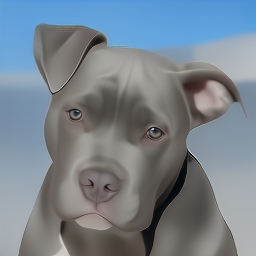}&
\includegraphics[scale=0.19]{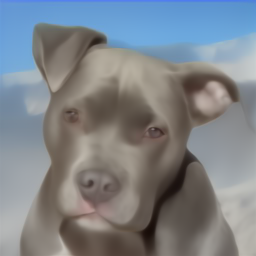}&
\includegraphics[scale=0.19]{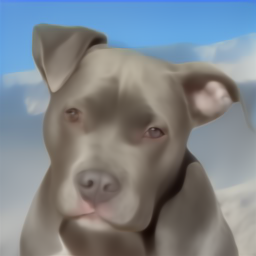}
\\
\includegraphics[scale=0.19]{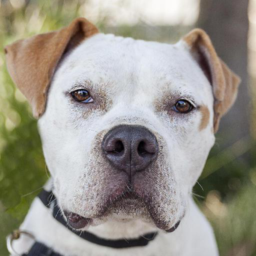}&
\includegraphics[scale=0.19]{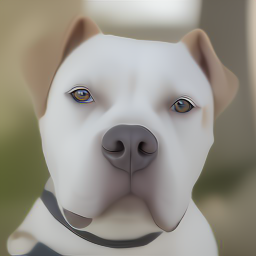}&
\includegraphics[scale=0.19]{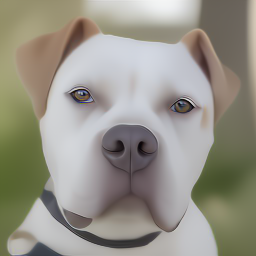}&
\includegraphics[scale=0.19]{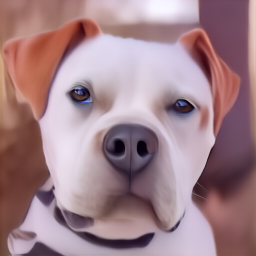}&
\includegraphics[scale=0.19]{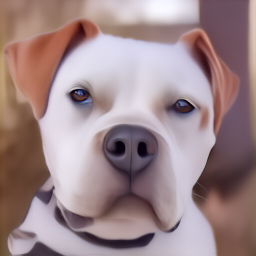}
\\
\includegraphics[scale=0.19]{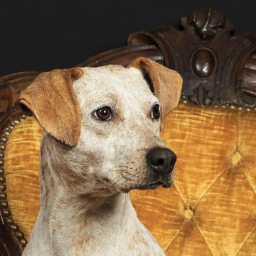}&
\includegraphics[scale=0.19]{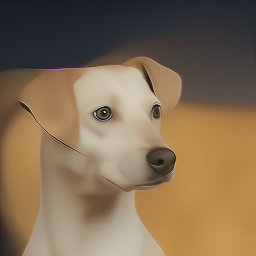}&
\includegraphics[scale=0.19]{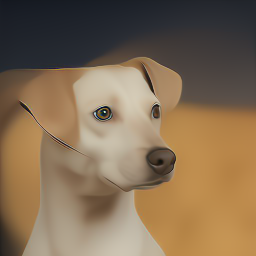}&
\includegraphics[scale=0.19]{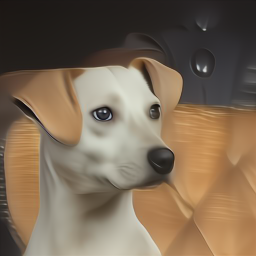}&
\includegraphics[scale=0.19]{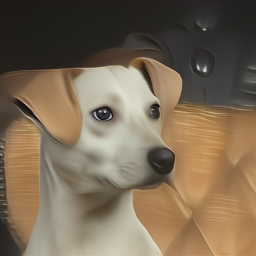}
\end{tabular}
\vspace{3mm}
\caption{Examples of generated images using the ST approach of our framework on AFHQ dataset.}
\label{fig:app_c}
\end{figure}

The task is animal category recognition into three domains of cat, dog, and wildlife, and the classifier is ResNet-18 with 99.73\% accuracy. Since we obtained a pretrained diffusion model on AFHQ-Dog dataset from~\cite{kim2022diffusionclip}, the diffusion model in our experiments could only generate dog images, and work well with the ST approach. Visual examples using the ST approach of our framework on AFHQ dataset are shown in Figure~\ref{fig:app_c}.

